\documentclass[10pt,journal]{IEEEtran}
\usepackage{amsmath,amssymb,amsfonts,amsthm}
\usepackage{array}
\usepackage[caption=false,font=footnotesize,labelfont=rm,textfont=rm]{subfig}
\usepackage{textcomp}
\usepackage{stfloats}
\usepackage{url}
\usepackage{cases}

\usepackage{multirow}
\usepackage{xcolor}
\usepackage{verbatim}
\usepackage{graphicx}

\usepackage{cite}
\usepackage{threeparttable}
\usepackage{booktabs}

\usepackage[ruled]{algorithm2e} 

\SetCommentSty{mycommfont}

\graphicspath{{./fig/}}

\def\BibTeX{{\rm B\kern-.05em{\sc i\kern-.025em b}\kern-.08em
    T\kern-.1667em\lower.7ex\hbox{E}\kern-.125emX}}
\usepackage{balance}

\begin{document}
\bstctlcite{setting}

\title{Large Vision Model-Enhanced Digital Twin with Deep Reinforcement Learning for User Association and Load Balancing in Dynamic Wireless Networks}

\author{Zhenyu~Tao,~\IEEEmembership{Student Member,~IEEE},
        Wei~Xu,~\IEEEmembership{Fellow,~IEEE},
        and Xiaohu~You,~\IEEEmembership{Fellow,~IEEE}

\thanks{An earlier version of this paper was accepted by the 34th International Joint Conference on Artificial Intelligence (IJCAI 2025).}
\thanks{Z. Tao, W. Xu, and X. You are with the National Mobile Communications Research Lab, Southeast University, Nanjing 210096, China, and also with the Pervasive Communication Research Center, Purple Mountain Laboratories, Nanjing 211111, China (email: \{zhenyu\_tao, wxu, xhyu\}@seu.edu.cn).}
}


\maketitle


\begin{abstract}
Optimization of user association in a densely deployed cellular network is usually challenging and even more complicated due to the dynamic nature of user mobility and fluctuation in user counts. While deep reinforcement learning (DRL) emerges as a promising solution, its application in practice is hindered by high trial-and-error costs in real world and unsatisfactory physical network performance during training. Also, existing DRL-based user association methods are typically applicable to scenarios with a fixed number of users due to convergence and compatibility challenges. To address these limitations, we introduce a large vision model (LVM)-enhanced digital twin (DT) for wireless networks and propose a parallel DT-driven DRL method for user association and load balancing in networks with dynamic user counts, distribution, and mobility patterns. To construct this LVM-enhanced DT for DRL training, we develop a zero-shot generative user mobility model, named Map2Traj, based on the diffusion model. Map2Traj estimates user trajectory patterns and spatial distributions solely from street maps. DRL models undergo training in the DT environment, avoiding direct interactions with physical networks. To enhance the generalization ability of DRL models for dynamic scenarios, a parallel DT framework is further established to alleviate strong correlation and non-stationarity in single-environment training and improve training efficiency. Numerical results show that the developed LVM-enhanced DT achieves closely comparable training efficacy to the real environment, and the proposed parallel DT framework even outperforms the single real-world environment in DRL training with nearly 20\% gain in terms of cell-edge user performance.
\end{abstract}


\section{Introduction}

\IEEEPARstart{T}{he} long-term evolution of c ellular networks, with increasing heterogeneity, density, and multi-band usage, has significantly raised the complexity of wireless networks \cite{10183795}. In the context of user association and load balancing, traditional strategies that focus solely on maximizing the signal-to-interference-plus-noise ratio (SINR) often result in unbalanced load distribution across bands and cells, impairing the overall network throughput and coverage \cite{10298039}. The vast scale and complexity of wireless networks, coupled with the dynamics of user mobility and varying user densities, render the optimization of user association strategy inherently non-linear and computationally demanding. This is particularly true in cellular-based vehicle-to-everything (C-V2X) networks, where vehicle users exhibit high spatial dynamics and stringent demands for both reliability and throughput. Deep reinforcement learning (DRL) has emerged as a promising solution to address extensive state spaces and nonconvex optimization problems, gaining widespread adoption in user association tasks \cite{10298039,9789983,8796358,tao2023DTAC}.

Due to the scarcity of effective virtual training environments, current DRL-based user association approaches are supposed to be directly trained in real-world wireless networks in applications. However, real-world training is fraught with challenges, including prohibitive trial-and-error costs and poor network performance before convergence \cite{9372298}. Furthermore, training within a single real-world environment that varies over time makes it difficult for DRL agents to explore a globally optimal policy. Consequently, existing DRL-based user association methods typically support a fixed number of users. In this context, constructing a virtual training environment that closely mimics the real-world scenario, also known as a digital twin (DT), becomes crucial for applying DRL in wireless networks and ensuring both the efficacy of DRL and the physical network performance during training \cite{tao2023wireless}. 

In order to faithfully replicate the real-world wireless network, a DT of a wireless network should integrate not only a precise wireless channel model within the area of interest, commonly referred to as the radio map \cite{8648450}, but also an accurate user mobility model due to its significant influence on the distribution and transition pattern of channel states and base station (BS) loads. While the modeling of wireless channel has been extensively studied and can be implemented deterministically through measurement and ray-tracing approaches \cite{9237116}, the modeling of user mobility still lacks effective, efficient, and privacy-preserving methodologies \cite{8673556}.
Specifically, most existing network optimization works \cite{9789983,8796358} employed random mobility models, typically the random waypoint model \cite{Johnson1996} and the Gauss Markov model \cite{752157}, to represent user mobility patterns. While these models can partially simulate random user movements, their adherence to specific distributions can lead to a significant mismatch with real-world trajectory patterns and spatial distribution. This mismatch usually results in considerable performance degradation of DRL models in practice. Although this issue can be alleviated to some extent by employing trace-based mobility models, real user trajectories are often inaccessible due to data acquisition costs and privacy concerns \cite{8673556}. 

The rapid advancement of artificial intelligence (AI) has enabled transformative applications across diverse domains, driven by breakthroughs in large-AI models. Large language models (LLMs) like GPT-4 \cite{achiam2023gpt} and DeepSeek \cite{liu2024deepseek} excel in text-based tasks, while large vision models (LVMs) such as stable diffusion \cite{rombach2022high} and Sora \cite{liu2024sora} demonstrate remarkable capabilities in visual content generation.
Recently, LVMs, particularly generative adversarial networks (GANs) \cite{goodfellow2020generative} and diffusion models \cite{NEURIPS2020_4c5bcfec}, have been applied to trajectory generation tasks and achieved promising outcomes \cite{9338370,trajgen,tstrajgen,DiffTraj,SynMob}. However, existing methods necessitate a significant amount of real trajectories to learn a trajectory distribution in a given area, leading to a paradox. That is, AI models struggle to generate lifelike and useful trajectories without ample real trajectories. Conversely, when ample real data is available, AI models tend to be unnecessary, as trace-based mobility models can be readily constructed. Therefore, existing trajectory generation methods are limited to functioning as data augmentation and privacy protection tools for trace-based mobility models, rather than serving as genuine generative user mobility models.

To address all these issues, in this paper, we introduce an LVM-enhanced DT for wireless networks featuring a novel zero-shot generative user mobility model, along with a parallel DT-driven DRL framework for user association and load balancing in dynamic wireless networks. The contributions of this work are summarized as follows.

\begin{itemize}

\item{We develop a distributed DRL method to achieve user association and load balancing in dynamic wireless networks. This approach includes a distributed association strategy to accommodate varying user numbers, and an enhanced DRL model to accelerate the convergence.
}

\item{We propose an LVM-enhanced DT construction technique for physical wireless networks, featuring a zero-shot generative user mobility model, named Map2Traj. This user mobility model is based on LVM, specifically the diffusion model, and generates lifelike trajectories solely from street maps. To our knowledge, this is the first work to achieve zero-shot trajectory generation, especially for mobile terminals in wireless networks.
} 

\item{A parallel DT framework is further established to enhance the generalization ability of DRL models. We analyze the limitations of single real-world environment DRL training, especially for user association tasks in dynamic wireless networks, and address this with the proposed parallel framework.}

\item{The efficacy of the proposed method is validated through real-world vehicle trajectory datasets and comprehensive experiments. The evaluation involves assessing the fidelity of the LVM-enhanced DT, its reliability in DRL training, and the performance of the parallel DT-driven DRL method for user association and load balancing.
}


\end{itemize}

\section{Related Work}

\subsection{DRL-based User Association and Load Balancing}
Traditional strategies that maximize SINR in user association result in unbalanced load distribution and consequently hinder the overall performance of wireless networks. DRL was first introduced in \cite{8796358} to solve the non-convex user association problems, maximizing long-term overall network utility through multi-agent DRL. Different from \cite{8796358} that uses global network information, the authors in \cite{9127161} trained independent agents on local observation, achieving scalable and flexible user association in a millimeter-wave network. Taking user mobility and handovers into consideration, studies in \cite{9789983} and \cite{9685781} employed a multi-armed bandit and a Q-learning framework, respectively, with the purpose of reducing handovers and improving throughput. 

However, due to the lack of effective virtual training environments, all these existing DRL-based approaches have to be trained in real-world scenarios, which raises concerns about trial-and-error costs and poor network performance before convergence. Moreover, constrained by the single real-world training environment, the DRL agents can hardly get enough generalization ability to cope with dynamic changes over long-time scales, for example, the variation of user numbers in the environment. Although some works have the potential to be applied in scenarios with different user densities, all existing DRL methods are set to be trained in a scenario with fixed user numbers, without the ability to adapt to dynamic user numbers.

\subsection{DT-driven Network Optimization}
To address the excessive convergence delay of DRL and its potential detriment to physical network performance, the DT technique has been widely adopted in learning-based network optimization. For integrated data and energy transfer tasks in cell-free networks, DTs have been proven effective in addressing performance fluctuations during DRL training and convergence acceleration \cite{10078846,10234388}. Additionally, in \cite{10345669}, the authors applied DTs to space-air-ground integrated networks, proposing a multi-agent DRL method designed to minimize latency, expand coverage, and reduce energy consumption. In scenarios involving edge computing offloading, studies such as \cite{9795902,10551488,10319104,9174795} leveraged both DT and DRL to improve the offloading performance across various metrics, including energy consumption, offloading efficiency, and latency.

Despite the diversity in tasks and scenarios, these studies share a common requirement, that is, the modeling of user mobility to create DTs of wireless networks. In \cite{10078846}, a simple static user location model was assumed, neglecting their mobility. The random waypoint (RWP) model, which involves moving to a randomly selected destination and then choosing the next one arbitrarily, was adopted by \cite{10234388} and \cite{10551488}. On the other hand, studies in \cite{10345669} and \cite{9795902} employed the Gauss Markov model (GM), characterized by using a stochastic process to model changes in user velocity and direction. Both studies in \cite{10319104} and \cite{9174795} employed the real-world T-Drive vehicle data set \cite{1869807} to establish the aforementioned trace-based mobility model. 

However, neither static assumptions nor random mobility models, including RWP and GM, can accurately capture the complex patterns of real-world user mobility, potentially compromising the effectiveness of DRL once deployed. While real data-based models offer a more precise representation, they come with the drawbacks of high data acquisition costs, privacy concerns, and limited transferability to other regions.

\subsection{LVM-based Trajectory Generation} 

Initially, generative AI and LVMs were introduced to trajectory generation primarily for synthesizing mobility data, thereby safeguarding the privacy of data providers. Liu et al. \cite{liu2018trajgans} first discussed the possibility of using GANs for trajectory generation, albeit without providing a detailed approach. After that, TrajGAIL \cite{9338370} was developed by employing generative adversarial imitation learning (GAIL), which combines both DRL and GAN, to generate trajectories through a series of next-location predictions. In \cite{trajgen}, another approach named TrajGen was developed by transforming trajectories into images and using a deep convolutional GAN (DCGAN) to generate virtual trajectory images. 
In \cite{tstrajgen}, TS-TrajGen was proposed to integrate GAN with the mobility analysis method, including the A* algorithm and mobility yaw reward, to enhance the model performance. 
DiffTraj in \cite{DiffTraj} applied a diffusion model to generate synthetic trajectories while preserving spatial-temporal features extracted from real trajectories. These studies have demonstrated commendable performance in generating privacy-preserving synthetic trajectories.

However, these methods fall short when it comes to zero-shot trajectory generation for unobserved new areas. Specifically, TrajGAIL \cite{9338370} simply samples actions from generated action probability distributions and constructs trajectories autoregressively, without the capacity to introduce data from new areas. Although TrajGen \cite{trajgen} used street map data to filter and calibrate generated trajectories through map matching \cite{mapmatching}, the generated trajectories adhered to the training set distribution, rather than that in new areas. Building further upon TrajGen, TS-TrajGen \cite{tstrajgen} utilized street maps to select and construct the best continuous trajectory with the A* algorithm. This process, however, remained confined to trajectories that adhere to the original distribution. Alternatively, in a conditional generation manner, DiffTraj \cite{DiffTraj} employed the diffusion model and incorporated prior knowledge of trip data, such as the travel time, average speed, and distance. While this complementary knowledge does improve generation performance, it does not enable the transfer of trajectory generation to new areas.


\section{System Model} 
\label{systemmodel}

Consider a downlink cellular network with $K$ distinct carrier frequency bands. Let $\mathcal{B}_k$ represent the set of all BSs operating on the $k$-th band, where $k=1,2,\dots,K$, and let
$\mathcal{U}$ denote the set of users. It is important to note that BSs are discriminated by both their physical location and the carrier frequency band they use. Thus, radio units on the same physical tower but operating on different frequency bands are considered co-located but distinct BSs. The set of all BSs is represented by $\mathcal{B} = \bigcup_{k=1}^K \mathcal{B}_k$.

We assume the user association is conducted on a larger timescale compared to the scheduling of time and frequency resources \cite{6497017}, and henceforth refer to the user association interval as a \textit{time slot}. During time slot $n$, the SINR for user $i\in\mathcal{U}$ from BS $j \in \mathcal{B}_k$ is given by
\begin{equation}
    \operatorname{SINR}_{i j}[n]=\frac{P_j g_{i j}[n]}{\sum_{j^{\prime} \in \mathcal{B}_k, j^{\prime} \neq j} P_{j^{\prime}} g_{i j^{\prime}}[n]+W_j \sigma^2},
\end{equation}
where $P_j$ represents the transmit power of BS $j$, $g_{i j}[n]$ denotes the channel gain between user $i$ and BS $j$, $W_j$ denotes the bandwidth available at BS $j$, which varies across different frequency bands, and $\sigma^2$ denotes the energy of thermal noise. The SINR is averaged within a time slot, eliminating the effects of small-scale fading, and thus the channel gain $g_{i j}[n]$ is only associated with antenna gains, pathloss, and shadow fading. Note that the channel gain can be obtained by users from reference signal received power (RSRP) measurements \cite{3gpp}. Then, the achievable rate between user $i$ and BS $j$ at time slot $n$ can be written as
\begin{equation}
    c_{i j}[n]=W_j \log _2\left(1+\operatorname{SINR}_{i j}[n]\right).
\end{equation}
Considering that handover between BSs interrupts normal data transmission, we define $T_{\mathrm{HO},i}[n]$ and $T_s$ respectively as the handover interruption time for user $i$ and the slot duration. $T_{\mathrm{HO},i}[n]$ is $0$ when the user stays connected to the original BS, and a positive value when the user successfully switches to another BS. Importantly, a handover attempt may not always succeed, and its failure could lead to an increased interruption time, as the user needs to reestablish its connection via the radio resource control (RRC) protocol. Then, the service rate for user $i\in\mathcal{U}$ from BS $j \in \mathcal{B}_k$ over time slot $n$ is expressed as
\begin{equation}
    r_{i j}[n]=(c_{i j}[n]/\ell_j[n])(1-T_{\mathrm{HO},i}[n]/T_s)
\end{equation}
where $\ell_j[n]$ denotes the number of users served by BS $j$, i.e., the load on BS $j$, at time slot $n$. 

The user association decision is made according to some specific policy at the beginning of each time slot $n$. For an arbitrary user $i$, the decision $\mathbf{x}_{i}[n]$ at time slot $n$ encompasses a set of association indicators, represented as $\left\{x_{i j}[n] | j \in \mathcal{B}\right\}$, where
\begin{equation}
    x_{i j}[n]= \begin{cases}1, & \text { user } i \text { associates to BS } j \text { in time slot } n, \\ 0, & \text { otherwise. }\end{cases} \label{maxeq1}\notag
\end{equation}
Since each user can only be assigned to one BS, we have
\begin{equation}
\sum_{j \in \mathcal{B}}\nolimits x_{i j}[n]=1, \quad \forall i \in \mathcal{U}. \label{maxeq2}
\end{equation}

The logarithmic utility function $\log(r_{i j}[n])$ is adopted as the optimization target to balance the sum-rate and fairness, which naturally achieves load-balancing \cite{srikant2014communication}. Now, the optimization task, i.e., the network utility maximization problem, can be formulated as
\begin{subequations}
\label{problem}
\begin{align}
 \underset{\left\{x_{i j}[n]\right\}}{\operatorname{maximize}}\ &\sum_{n=1}^T \sum_{j \in \mathcal{B}} \sum_{i \in \mathcal{U}} x_{i j}[n] \log \left(\frac{c_{i j}[n]}{\ell_j[n]} \cdot\left(1-\frac{T_{\mathrm{HO},i}[n]}{T_s}\right)\right) \\
 \text {subject to}\ & \text{(\ref{maxeq1}), (\ref{maxeq2}).}
\end{align}
\end{subequations}

\section{Distributed DRL Method for User Association}\label{DRLmethod}
\begin{figure}[!t]
\centering
\includegraphics[width=0.5\textwidth]{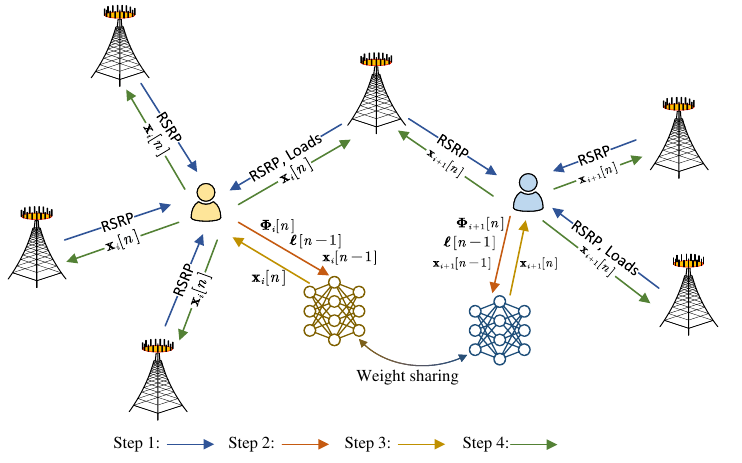}

\caption{Schematic of distributed DRL for user association}
\label{scenario}
\end{figure}
The prohibitive computational complexity and inability to predict SINR and handover events limit the application of integer linear programming or searching-based methods for solving the optimization problem in (\ref{problem}). Moreover, the potential variation in user number results in fluctuation in dimensions of the state and action for centralized DRL methods, thus hindering their application in this problem. We develop a distributed multi-agent DRL method to cope with these issues. In our method, distributed DRL agents with shared parameters are assigned to users to make association decisions. The number of agents adjusts with the number of users, while their dimensions of state and action remain constant. The problem in (\ref{problem}) is formulated within a Markov decision process (MDP) framework, characterized by the 5-tuple $\langle\mathcal{S}, \mathcal{A}, \mathcal{T}, \mathcal{R}, \gamma\rangle$, which represent the state space, action space, state transition probability, reward, and discount factor, respectively. For the sake of clarity and without loss of generality, we select an arbitrary user $i$ to illustrate this MDP.

\textbf{State:}
Denote the SINRs between user $i$ and BSs as a vector $\boldsymbol{\Phi}_i[n]$, whose $j$-th element is $\operatorname{SINR}_{i j}[n]$. The loads on BSs are represented by vector $\boldsymbol{\ell}[n]$, consisting of all $\ell_j[n]$ for $j\in\mathcal{B}$. Then, the state of the DRL agent is defined by
\begin{equation}
    \mathbf{s}_i[n] = (\boldsymbol{\Phi}_i[n],\ \boldsymbol{l}[n-1],\ \mathbf{x}_{i}[n-1]), \label{state}
\end{equation}
where $\mathbf{x}_{i}[n]$ is the association decision in the last slot that helps the agent assess whether there will be a handover. $\boldsymbol{\Phi}_i[n]$, $\boldsymbol{\ell}[n-1]$, and $\mathbf{x}_{i}[n-1]$ are all $|\mathcal{B}|$-length vectors, where $|\mathcal{B}|$ denotes the cardinality of set $\mathcal{B}$. The variation of user number in the environment only affects the value of $\boldsymbol{\ell}[n-1]$, without changing the dimension of $\mathbf{s}_i[n]$.

\textbf{Action:} The action, denoted by $\mathbf{a}_i[n]$, is defined as the association decision $\mathbf{x}_{i}[n]$ for user $i$, which is also a fixed $|\mathcal{B}|$-length vector. The $i$-th DRL agent makes this decision based on the current state $\mathbf{s}_i[n]$.

The DRL agent takes partial observation $\mathbf{s}_i[n]$ as input, infers other users' information through the value and changes of $\boldsymbol{l}[n-1]$, and outputs $\mathbf{a}_i[n]$ to select a BS. As depicted in Fig.~\ref{scenario}, the entire association procedure consists of four steps. First, the user measures RSRP from all BSs in the environment and gets all BSs' loads in the previous time slot from the currently connected BS. Second, the user calculates $\boldsymbol{\Phi}_i[n]$ based on RSRP, and transfers it to the agent, together with $\boldsymbol{\ell}[n-1]$ and $\mathbf{x}_{i}[n-1]$. Then, the agent outputs the association decision $\mathbf{x}_{i}[n]$ to the user based on the transferred information, by which the user can finally connect to the designated BS.

\textbf{State transition:}
The state transition probability is denoted by $\mathbb{P}(\mathbf{s}_i[n+1]|\mathbf{s}_i[n],\mathbf{a}_i[n])$, indicating the probability of transitioning into state $\mathbf{s}_i[n+1]$ when action $\mathbf{a}_i[n]$ is made at state $\mathbf{s}_i[n]$, where $\mathbb{P}(\cdot)$ represents the probability of an event.

\textbf{Reward:}
To balance the convergence rate and overall network utility, we define the reward for the $i$-th agent as a function of both the individual utility of user $i$ and the aggregate utility of all users, as follows:
\begin{align}
&R_i[n] = \alpha \sum\nolimits_{j \in \mathcal{B}} x_{i j}[n] \log \left(r_{i j}[n]\right)  \notag \\
 +&\frac{1-\alpha}{|\mathcal{B}|} \sum\nolimits_{j \in \mathcal{B}} \sum\nolimits_{k \in \mathcal{U}} x_{k j}[n] \log \left(r_{k j}[n]\right), \ \alpha\in \left[0,1\right].     \label{reward}
\end{align}
The $\alpha$ is a hyperparameter to balance between the convergence rate and overall performance. Notice that $r_{i j}[n]$ in (\ref{reward}) is derived from $\operatorname{SINR}_{i j}[n]$ (part of $\boldsymbol{\Phi}_i[n]$ from $\mathbf{s}_i[n]$) and $\ell_j[n]$ (part of $\boldsymbol{\ell}[n]$ from $\mathbf{s}_i[n+1]$). In addition, $x_{i j}[n]$ in (\ref{reward}) is an element of the action vector $\mathbf{a}_i[n]$. Therefore, given the transmit power $P_j$ and bandwidth $W_j$ for each $j \in \mathcal{B}$, as well as handover interruption time $T_{\mathrm{HO},i}$, the reward for the $i$-th DRL agent can be written in a function as 
\begin{equation}
    R_i[n] = R(\mathbf{s}_i[n],\ \mathbf{a}_i[n],\ \mathbf{s}_i[n+1]).
\end{equation}


Now that the optimization target of DRL is readily defined as follows
\begin{equation}
    \underset{\pi_{\boldsymbol{\theta}}}{\operatorname{max}} \underset{\substack{
    i \in \mathcal{U},\ \mathbf{s}_i[n] \sim \mathcal{D}\\
    \mathbf{a}_i[n] \sim \pi_{\boldsymbol{\theta}}(\cdot|\mathbf{s}_i[n])\\
    \mathbf{s}_i[n+1] \sim \mathbb{P}(\cdot|\mathbf{s}_i[n],\mathbf{a}_i[n])}
    }{\mathbb{E}} \left[\sum_{n=0}^\infty \gamma^n R_i[n]\right] ,\label{maximize}
\end{equation}
where $\mathcal{D}$ is the distribution over state space $\mathcal{S}$ in physical wireless networks, and $\pi_{\boldsymbol{\theta}}$ is the parameterized stochastic policy which maps states into action probabilities.

Since all DRL agents share the same parameters and weights, training samples $(\mathbf{s}_i[n],\ \mathbf{a}_i[n],\ \mathbf{s}_i[n+1],\ R_i[n]), \ \forall i \in \mathcal{U}$ are not distinguished by subscripts and are uniformly used for training. The training strategy here follows the state-of-the-art proximal policy optimization (PPO) algorithm \cite{schulman2017proximal}, with several modifications to the structure of the actor network. Specifically, we integrate a long short-term memory (LSTM) layer in the neural network to capture the dynamics of states and incorporate memory capability. In addition, the input $\boldsymbol{\Phi}_i[n]$ is utilized to construct an action mask that selects top-$N$ BSs with the strongest signal, thereby assisting the convergence.
\section{Parallel DT-driven DRL Method} 

\subsection{Construction of the LVM-enhanced DT} %
\label{constructDT}
To address the high trial-and-error costs, poor network performance before convergence, and limitations of the single training environment in traditional DRL methods, it is crucial to develop a faithful DT environment for effective DRL training. By aligning the number of BSs and users, the state space $\mathcal{S}$ and action space $\mathcal{A}$ can be mirrored from the real environment. The training process for DRL agents in such a DT environment can be expressed in a way similar to (\ref{maximize}) as
\begin{equation}
    \underset{\pi_{\boldsymbol{\theta}}}{\operatorname{max}} \underset{\substack{
    i \in \mathcal{U},\ \mathbf{s}_i[n] \sim \mathcal{D^\prime}\\
    \mathbf{a}_i[n] \sim \pi_{\boldsymbol{\theta}}(\cdot|\mathbf{s}_i[n])\\
    \mathbf{s}_i[n+1] \sim \mathbb{P^\prime}(\cdot|\mathbf{s}_i[n],\mathbf{a}_i[n])}
    }{\mathbb{E}} \left[\sum_{n=0}^\infty \gamma^n R_i[n]\right],
\end{equation}
where $\mathcal{D^\prime}$ denotes the state distribution within the DT environment, and $\mathbb{P^\prime}$ symbolizes the state transition probability. Since the reward is contingent upon state and action, the main challenge is to approximate real-world state distribution $\mathcal{D}$ and transition probabilities $\mathbb{P}$ through $\mathcal{D^\prime}$ and $\mathbb{P^\prime}$.

To analyze the prerequisite for constructing $\mathcal{D^\prime}$ and $\mathbb{P^\prime}$, we first revisit the definition of state in (\ref{state}). Among the three elements of the state vector, $\mathbf{x}_{i}$ is determined by the policy $\pi_{\boldsymbol{\theta}}$ at the preceding time slot. Therefore, the distribution and transition probability of states are predominantly shaped by $\boldsymbol{\Phi}_{i}$ and $\boldsymbol{\ell}$. Here, $\boldsymbol{\Phi}_{i}$ represents the SINRs between a single user and BSs, which depend on the wireless channel and the user's location. Thus, the distribution of $\boldsymbol{\Phi}_{i}$ is determined by the channel and the \textit{spatial distribution} of the user, while the transition probability is decided by the channel and the evolution of the user's location, i.e., the \textit{trajectory pattern}. 

Similarly, the load vector $\boldsymbol{\ell}$ depends not only on the wireless channel and user location but also on the association policy $\pi_{\boldsymbol{\theta}}$.  The distribution and transition probabilities of $\boldsymbol{\ell}$ share similar determinants with $\boldsymbol{\Phi}_{i}$, with the additional influence of the association policy. Furthermore, the mobility factors for $\boldsymbol{\ell}$, including \textit{spatial distribution} and \textit{trajectory pattern}, extend from an individual user to all users within the environment. In summary, the wireless channel model and the user mobility model, encompassing both \textit{spatial distribution} and \textit{trajectory pattern} of users, are fundamental models for constructing a faithful DT environment.

While channel modeling has been extensively studied and can be implemented deterministically through measurement and ray-tracing methods \cite{9237116}, there are, thus far, few methods capable of efficiently modeling user mobility within a newly given area \cite{8673556}. The random mobility models can hardly capture the distribution and trajectory patterns of real-world users due to their adherence to the stochastic process. Trace-based methods and existing learning-based trajectory generation approaches ask for a large amount of data, constrained by the high data acquisition costs and privacy issues. Inspired by the zero-shot image generation \cite{pmlr-v139-ramesh21a}, which enables the creation of images from descriptions unseen during training, we devise a zero-shot trajectory generation approach to construct the generative user mobility model. 

\begin{figure}[!t]
\centering
\subfloat[Street map]{\includegraphics[width=0.14\textwidth]{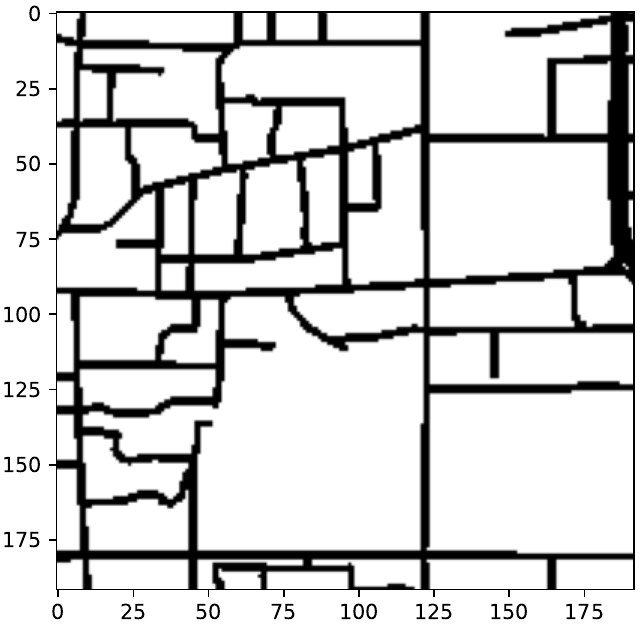}\label{Fig1.1}}
\subfloat[Trajectories]{\includegraphics[width=0.14\textwidth]{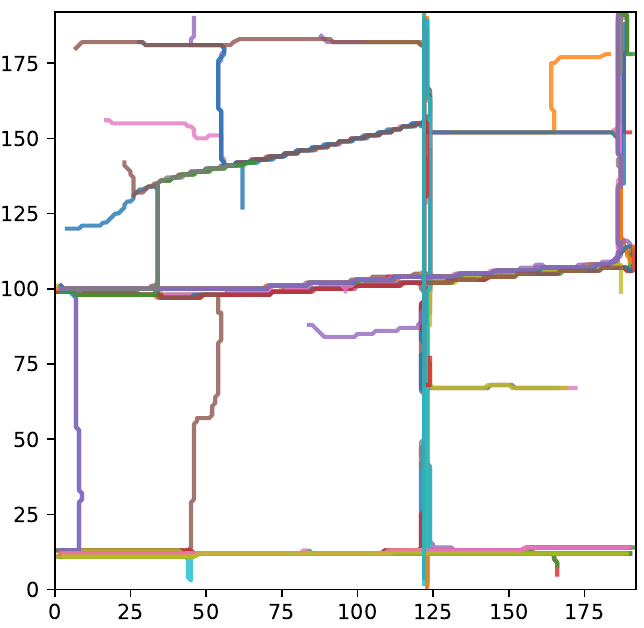}\label{Fig1.2}}
\subfloat[Heatmap]{\includegraphics[width=0.14\textwidth]{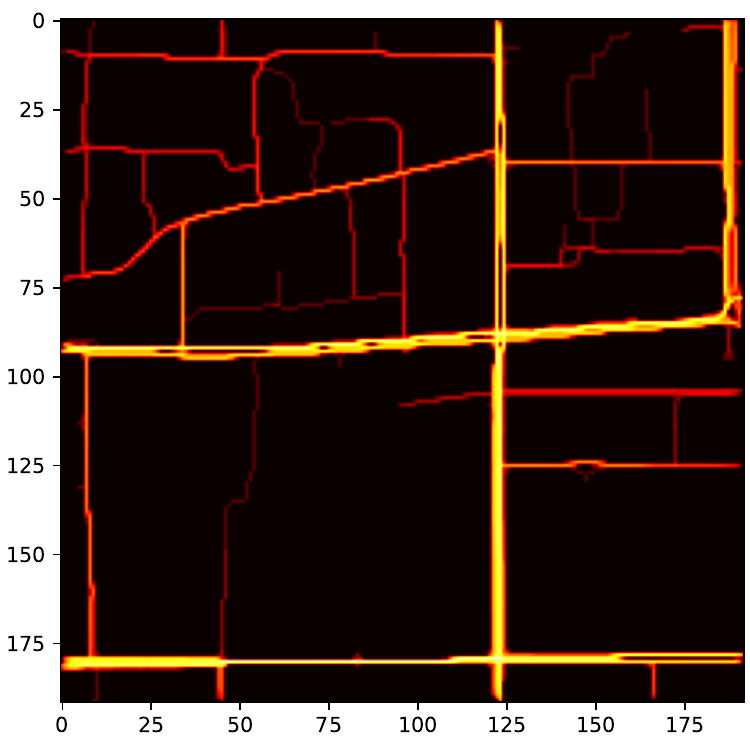}\label{Fig1.3}}
\caption{Correlation between street map and trajectories.}
\label{correlation}
\end{figure}

The key to zero-shot learning is to associate observed and unobserved objects through some form of auxiliary information, which encodes the inherent properties of objects \cite{xian2017zero}. The question arises: Can we apply zero-shot generation techniques to trajectory generation? More specifically, is there some form of data that is both readily accessible and closely related to real trajectories, akin to the relationship between texts and images? The answer lies in street maps, which are usually open-source and exhibit strong correlations with user trajectories and distribution, especially the high dynamic vehicle users, as depicted in Fig.~\ref{correlation}. Consequently, in this study, the objects in zero-shot learning are trajectories following different area-specific distributions, while the auxiliary information is the street map. By understanding the relationship between trajectories and maps through extensive training data, the proposed Map2Traj model generates lifelike trajectories for unobserved areas piloted by their street maps, that is, zero-shot trajectory generation.

Now we give a formal definition of this zero-shot trajectory generation task conditioned on street maps: Given a training set includes a set street maps $\mathcal{M} = \{\boldsymbol{m}^1,\boldsymbol{m}^2,\dots\}$ and the corresponding set of real-world trajectories $\mathcal{T} =\{\mathcal{L}^1,\mathcal{L}^2,\dots\}$, where each $\mathcal{L}^x = \{\boldsymbol{l}^{x,1},\boldsymbol{l}^{x,2},\dots\}$ is a set of real trajectories within the area of street map $\boldsymbol{m}^x$. The objective of zero-shot trajectory generation is to develop a generative model trained on this training set $\langle \mathcal{M}, \mathcal{T} \rangle$. With an unobserved street map $\boldsymbol{m}^o \notin \mathcal{M}$, this developed model should be capable of generating synthetic trajectories such that: 
\begin{itemize}
\item{Each generated trajectory closely resembles some real trajectory $\boldsymbol{l} \in \mathcal{L}^o$ in terms of \textit{trajectory pattern}, where $\mathcal{L}^o = \{\boldsymbol{l}^{o,1},\boldsymbol{l}^{o,2},\dots\}$ is a set of real trajectories within this unobserved area.
}
\item{The set of generative trajectories matches well with the real trajectory set $\mathcal{L}^o$ in terms of \textit{spatial distribution}.}
\end{itemize}
The high fidelity of generated trajectories in \textit{spatial distribution} and \textit{trajectory pattern} would guarantee faithful $\mathcal{D^\prime}$ and $\mathbb{P^\prime}$ in the constructed DT environments.

In this study, trajectories $\boldsymbol{l}$ and street maps $\boldsymbol{m}$ are both processed into 192$\times$192 binary images. Although a single binary image can effectively convey the spatial layout of a street map, it falls short in depicting distinct characteristics of various road types. In the OpenStreetMap dataset, roads are tagged with attributes such as \textit{Trunk}, \textit{Primary}, and \textit{Residential}. To exploit these attributes, we categorize roads into multiple groups, create binary images for each group, and merge them into a multi-channel binary image.

As a tractable solution for the zero-shot trajectory generation task, our proposed Map2Traj approach exploits the LVM, specifically the diffusion model \cite{NEURIPS2020_4c5bcfec}, to generate user trajectories. It consists of a forward diffusion process and a reverse diffusion process (denoising). 

\subsubsection{Forward Diffusion Process}
The forward diffusion process in Map2Traj adheres to the standard paradigm, which involves a Markovian process that iteratively adds Gaussian noise $\mathcal{N}(\cdot)$ to a real trajectory data $\boldsymbol{l}_0 \equiv \boldsymbol{l}$ over $T$ time steps:

\begin{align}
& q\left(\boldsymbol{l}_{t+1} | \boldsymbol{l}_t\right)=\mathcal{N}\left(\boldsymbol{l}_{t+1} ;  \sqrt{\alpha_t} \boldsymbol{l}_{t},\left(1-\alpha_t\right) \mathbf{I}\right), \\
& q\left(\boldsymbol{l}_{1: T} | \boldsymbol{l}_0\right)=\prod\nolimits_{t=1}^T q\left(\boldsymbol{l}_t | \boldsymbol{l}_{t-1}\right),
\end{align}
where $\alpha_t \text{ for } t=1,2,\dots,T$ are hyper-parameters of the noising process, $\mathcal{N}(x;\mu,\sigma)$ represents the normal distribution of mean $\mu$ and covariance $\sigma$ that produces $x$, and $\mathbf{I}$ denotes the identity matrix. This forward process noises a real trajectory $\boldsymbol{l}_0$ into $\boldsymbol{l}_T$, making it indistinguishable from Gaussian noise at the $T$-th step. Using the properties of Gaussian distributions, the forward process can also be marginalized to directly obtain the noisy trajectory at the $t$-th step. That is,
\begin{align}
q\left(\boldsymbol{l}_t | \boldsymbol{l}_0\right)=\mathcal{N}\left(\boldsymbol{l}_t ;  \sqrt{\gamma_t} \boldsymbol{l}_0,\left(1-\gamma_t\right) \mathbf{I}\right), \label{eq3}
\end{align}
where $\gamma_t=\prod\nolimits_{i=1}^t \alpha_i$.
Additionally, the parameterization of the Gaussian distribution of the forward process allows a closed-form formulation of the posterior distribution of $\boldsymbol{l}_{t-1}$ conditioned on $\boldsymbol{l}_0$ and $ \boldsymbol{l}_t$. It follows
\begin{align}
q\left(\boldsymbol{l}_{t-1} | \boldsymbol{l}_t,  \boldsymbol{l}_0\right)=\mathcal{N}\left(\boldsymbol{l}_{t-1} ; \boldsymbol{\mu}, \sigma^2 \mathbf{I}\right), \label{eq4}
\end{align}
where 
\begin{align}
    \boldsymbol{\mu}&=\frac{\sqrt{\gamma_{t-1}}\left(1-\alpha_t\right)}{1-\gamma_t} \boldsymbol{l}_0+\frac{\sqrt{\alpha_t}\left(1-\gamma_{t-1}\right)}{1-\gamma_t} \boldsymbol{l}_t, \label{eq4.1}\\  
    \sigma^2&=\frac{\left(1-\gamma_{t-1}\right)\left(1-\alpha_t\right)}{1-\gamma_t}.
\end{align}

\subsubsection{Reverse Diffusion Process}
\begin{figure}[!t]
\centering
\includegraphics[width=0.5\textwidth]{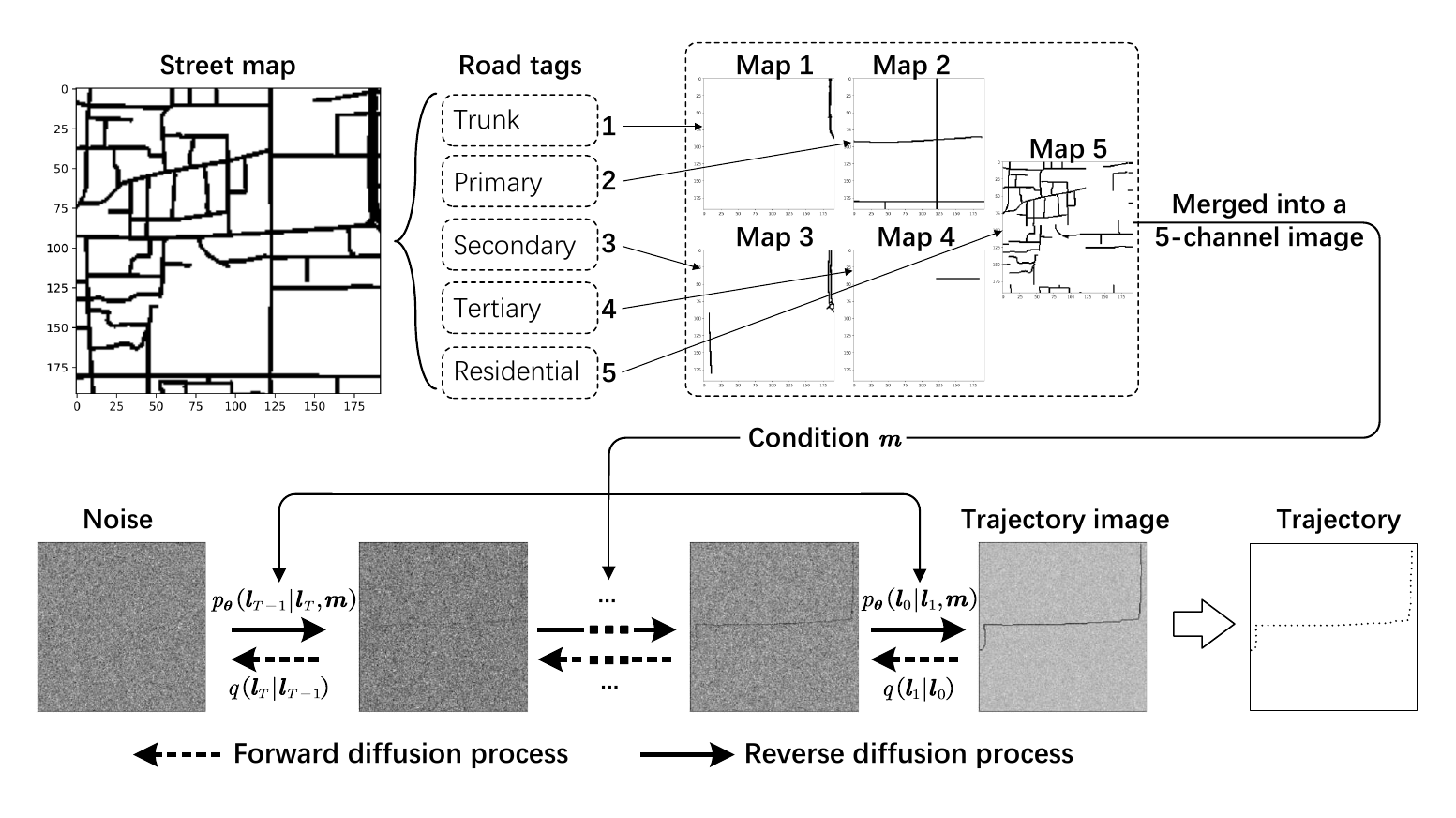}
\caption{Forward and reverse diffusion process in Map2Traj.}
\label{diffusion}
\end{figure}

The reverse process in the original diffusion model converts samples from a standard Gaussian distribution into outputs that follow a single target distribution. It is described by a parameterized probability distribution over the $T$ steps, defined as
\begin{equation}
    p_{\boldsymbol{\theta}}\left(\boldsymbol{l}_{0: T}\right)=p\left(\boldsymbol{l}_T\right) \prod\nolimits_{t=1}^T p_{\boldsymbol{\theta}}\left(\boldsymbol{l}_{t-1} | \boldsymbol{l}_t\right). \label{18}
\end{equation}
where
\begin{equation}
    p(\boldsymbol{l}_T) = \mathcal{N}(\boldsymbol{l}_T;0, \mathbf{I}).
\end{equation}
However, trajectories generated in this way always follow a single target distribution and cannot be applied to unobserved new regions. To address this, Map2Traj incorporates trajectories from various areas and uses the corresponding street maps as conditional inputs, extending the original single target distribution, $p_{\boldsymbol{\theta}}(\boldsymbol{l}_{0: T})$, into multiple area-specific target distributions, $p_{\boldsymbol{\theta}}(\boldsymbol{l}_{0: T} | \boldsymbol{m})$, where each distribution corresponds to a distinct street map $\boldsymbol{m}$. Then (\ref{18}) is reformulated as
\begin{equation}
    p_{\boldsymbol{\theta}}\left(\boldsymbol{l}_{0: T}| \boldsymbol{m}\right)=p\left(\boldsymbol{l}_T\right) \prod\nolimits_{t=1}^T p_{\boldsymbol{\theta}}\left(\boldsymbol{l}_{t-1} | \boldsymbol{l}_t , \boldsymbol{m}\right). \label{denoise}
\end{equation}
This approach allows Map2Traj to estimate the trajectory distribution from an unobserved street map and generate synthetic trajectories that conform to this distribution through sampling. The entire process is illustrated in Fig. \ref{diffusion}. 

\subsubsection{Training of Map2Traj}
Given a noisy trajectory $\boldsymbol{l}_t$ sampled from (\ref{eq3}), we have
\begin{align}
\boldsymbol{l}_t=\sqrt{\gamma_t} \boldsymbol{l}_0+\sqrt{1-\gamma_t} \boldsymbol{\epsilon},\ \ \boldsymbol{\epsilon} \sim \mathcal{N}(\mathbf{0}, \mathbf{I})\label{eq5}
\end{align}
where the goal is to recover the target trajectory $\boldsymbol{l}_0$. A neural network, denoted by $f_{\boldsymbol{\theta}}(\boldsymbol{m}, \boldsymbol{l}_t, t)$, is utilized to predict the noise component in $\boldsymbol{l}_t$, conditioned on the street map $\boldsymbol{m}$, the noisy trajectory $\boldsymbol{l}_t$, and the noise level indicated by the time step $t$. The training of Map2Traj is conducted by minimizing the mean squared error loss. That is,
\begin{align}
\min\nolimits_{\boldsymbol{\theta}} \ \mathbb{E}_{\boldsymbol{m},\boldsymbol{l},t,\boldsymbol{\epsilon}}\|f_{\boldsymbol{\theta}}(\boldsymbol{m}, \sqrt{\gamma_t} \boldsymbol{l}_0+\sqrt{1-\gamma_t} \boldsymbol{\epsilon}, t)-\boldsymbol{\epsilon}\|^2.
\end{align}
\subsubsection{Inference of Map2Traj}
The sampling process starts with pure Gaussian noise $\boldsymbol{l}_T$, followed by $T$ refinement steps. Given any noisy trajectory $\boldsymbol{l}_t$, we can approximate the target trajectory by rearranging terms in (\ref{eq5}) as
\begin{equation}
    \hat{\boldsymbol{l}}_0=\frac{1}{\sqrt{\gamma_t}}\big(\boldsymbol{l}_t-\sqrt{1-\gamma_t} f_{\boldsymbol{\theta}}\left(\boldsymbol{m}, \boldsymbol{l}_t, t\right)\big).
\end{equation}
Substituting estimate $\hat{\boldsymbol{l}}_0$ into (\ref{eq4}) and (\ref{eq4.1}), we parameterize the mean of $p_{\boldsymbol{\theta}}\left(\boldsymbol{l}_{t-1} | \boldsymbol{l}_t , \boldsymbol{m}\right)$ in (\ref{denoise}) as
\begin{equation}
    \mu_{\boldsymbol{\theta}}\left(\boldsymbol{m}, \boldsymbol{l}_t, t\right)=\frac{1}{\sqrt{\alpha_t}}\left(\boldsymbol{l}_t-\frac{1-\alpha_t}{\sqrt{1-\gamma_t}} f_{\boldsymbol{\theta}}(\boldsymbol{m}, \boldsymbol{l}_t, t)\right).
\end{equation}
And the variance of $p_{\boldsymbol{\theta}}\left(\boldsymbol{l}_{t-1} | \boldsymbol{l}_t , \boldsymbol{m}\right)$ is approximated as $(1-\alpha_t)$, following the setting in \cite{NEURIPS2020_4c5bcfec}. With this parameterization, denoising at the $t$-th step is conducted through sampling as follows
\begin{equation}
\boldsymbol{l}_{t-1} \leftarrow \frac{1}{\sqrt{\alpha_t}}\left(\boldsymbol{l}_t-\frac{1-\alpha_t}{\sqrt{1-\gamma_t}} f_{\boldsymbol{\theta}}\left(\boldsymbol{m}, \boldsymbol{l}_t, t\right)\right)+\sqrt{1-\alpha_t} \boldsymbol{\epsilon}, \label{denoise2}
\end{equation}
where $\boldsymbol{\epsilon} \sim \mathcal{N}(\mathbf{0}, \mathbf{I})$. Through iterative sampling, Map2Traj generates trajectories that follow the estimated joint distribution $p_{\boldsymbol{\theta}}\left(\boldsymbol{l}_{0: T}| \boldsymbol{m}\right)$ from Gaussian noise.

\subsubsection{Post Processing of Trajectory Image}
To model the user mobility in wireless network optimization, the trajectory sequence needs to be reconstructed from the image. Since each image represents only one trajectory per user, we used a simple search algorithm to reconstruct the sequence from one end to the other, where each point represents the coordinate of the pixel center. An extra deep neural network is utilized to estimate the sojourn time at each point, as employed in \cite{endo2016classifying}, from which the speed and relative time stamp are derived.

\begin{figure}[!t]
\centering
\includegraphics[width=0.5\textwidth]{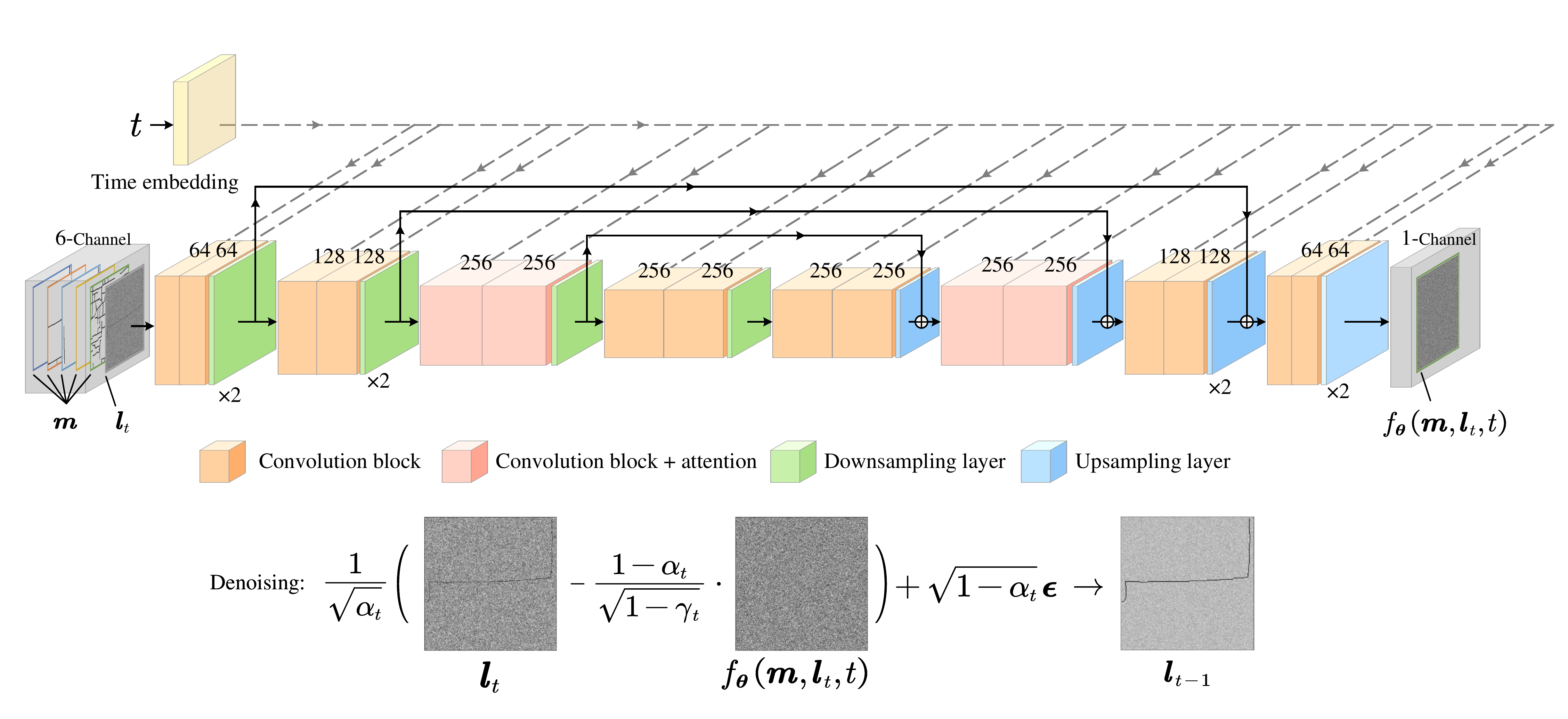}
\caption{The U-Net architecture in Map2Traj.}
\label{unet}
\end{figure}

\subsubsection{Neural Network Model in Map2Traj}
The neural network model utilized in Map2Traj, $f_{\boldsymbol{\theta}}(\boldsymbol{m}, \boldsymbol{l}_t, t)$, is based on a 192$\times$192 U-Net model \cite{ronneberger2015u}, with multiple modifications to improve its performance such as attention blocks \cite{oktay2018attention} and group normalization \cite{Wu_2018_ECCV}. A distinctive feature of Map2Traj is the incorporation of street map data through concatenation, allowing the model to be conditioned on the spatial information inherent in the maps. As illustrated in Fig.~\ref{unet}, street map $\boldsymbol{m}$ is concatenated with the noisy trajectory image $\boldsymbol{l}_t$, composing a multi-channel image as the input of the U-Net. Through multiple convolutions, attention mechanism, downsampling, and upsampling blocks, this multi-channel image is transformed into the prediction of the noise component in $\boldsymbol{l}_t$. The prediction, i.e., $f_{\boldsymbol{\theta}}(\boldsymbol{m}, \boldsymbol{l}_t, t)$, is represented as a single-channel image. The embedded time step $t$ is introduced to help the model assess the noise level in the trajectory image. We also provide an exemplified demonstration of the denoising process in (\ref{denoise2}) at the lower part of Fig.~\ref{unet}.

Now we can construct the DT for the wireless network in the area of interest, with the wireless channel model based on measurement and ray-tracing, as well as the user mobility model based on Map2Traj. We first place BSs in the DT with the same number, locations, power, and bandwidth as in the real network. This ensures that the state and action spaces remain statistically identical. Then, the wireless channel model is used to compute the RSRP between the BSs and users, and the user mobility model generates unique trajectories for each user in the environment. As users keep entering and leaving this area, the total number of users, $|\mathcal{U}|$, evolves over a long-time scale. 

The constructed DT operates in discrete time slots. At each time slot $n$, the user $i$ computes $\boldsymbol{\Phi}_i[n]$ based on the wireless channel model and its location, and composes $\mathbf{s}_i[n]$ using $\boldsymbol{l}[n-1]$ and $\mathbf{x}_{i}[n-1]$ from the previous time slot. The policy $\pi_{\boldsymbol{\theta}}$ then makes the association decision $\mathbf{a}_{i}[n]$ based on $\mathbf{s}_i[n]$ for each $i\in\mathcal{U}$, and the aggregate utility $R_i[n]$ is calculated. In the subsequent time slot, all users move into new locations according to their allocated trajectory, and the next state $\mathbf{s}_i[n+1]$ is determined accordingly. Each step in the DT generates $|\mathcal{U}|$ samples in the form of $(\mathbf{s}_i[n],\mathbf{a}_{i}[n],\mathbf{s}_i[n+1],R_i[n])$, which are utilized in the DRL training. 


\subsection{Parallel DT Framework}\label{parallelframework}
A straightforward approach to utilizing DT involves constructing a faithful DT environment, training DRL agents with collected data, and transferring the trained agents into the physical wireless network. Although this method addresses the issues of trial-and-error costs and poor physical network performance, it shares limitations with the single real-world environment.
 
\begin{figure}[!t]
\centering
\includegraphics[width=0.5\textwidth]{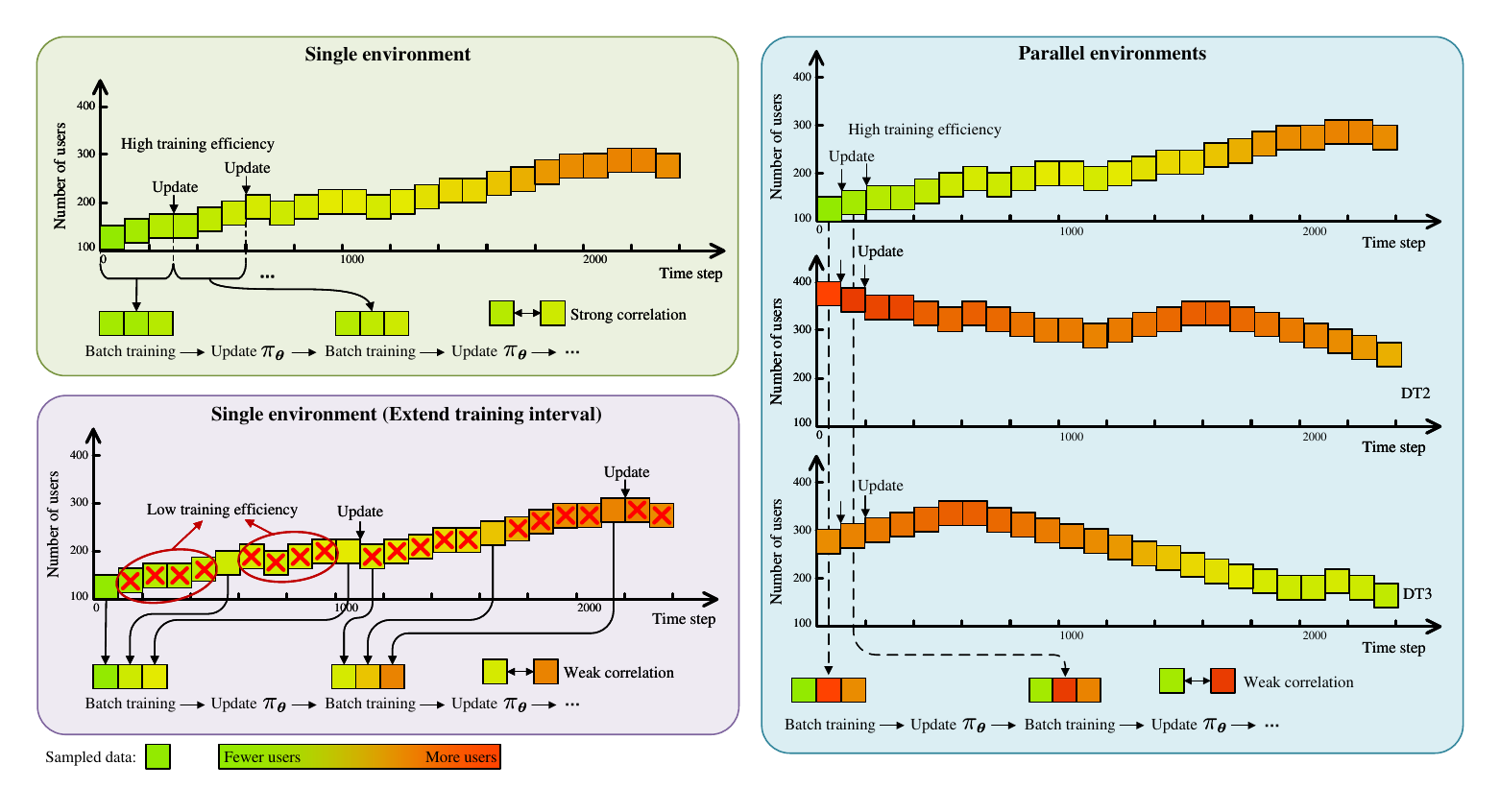}
\caption{Comparison between single and parallel environment training.}
\label{schematic}
\end{figure}

Specifically, the evolution of $|\mathcal{U}|$ in an environment occurs over a long-time scale (hundreds of time slots), while sample collection and batch training of $\pi_{\boldsymbol{\theta}}$ are performed over just one or a few time slots. As a result, the number of users in each training batch remains relatively stable or fluctuates slightly. This strong correlation of training samples on $|\mathcal{U}|$ would cause non-stationarity in DRL training, resulting in a suboptimal policy that focuses primarily on the current state, as schematized in Fig.~\ref{schematic}.

One potential solution to this issue in a single environment is to extend both the training interval and batch size, ensuring the batch training occurs only after $|\mathcal{U}|$ has sufficiently evolved. However, this approach would significantly reduce training efficiency. In multi-agent DRL systems, $\pi_{\boldsymbol{\theta}}$ can only undergo a limited number of gradient descents before being redeployed for further exploration. This is because the state and transition probabilities of the MDP for each agent are significantly affected by the actions of other agents. In the environment studied in this paper, this effect is observed in the strong relationship between actions of other agents under policy $\pi_{\boldsymbol{\theta}}$ and the distribution and transition probabilities of $\boldsymbol{l}[n]$ within the state $\mathbf{s}_i[n]$. While extending the training interval can mitigate this correlation, only a small portion of the collected samples would be effectively utilized for training, thus reducing the training efficiency. For instance, in our experiments, the traversal of the number of users from 100 to 400 requires 20,000 environment steps, generating $5\times10^6$ training samples, which are sufficient for DRL agents to converge. However, with an extended training interval, these samples would only support a few gradient descents, resulting in most of the samples being wasted. This low training efficiency is impractical for DRL training in both real and DT environments.

To address the dilemma between sample correlation and training efficiency, we develop a strategy using multiple parallel environments. Unlike the uncontrollable and non-reproducible real-world training environment, it is feasible to create multiple independent DT environments, each initialized with varying user densities. These environments help decorrelate the sample data, making the training process keep stationary, as agents in different DT environments experience a variety of states at each time step. This strategy with parallel architecture further exploits the potential of DT in DRL training, reducing sample correlation and improving training efficiency for DRL agents.

\begin{figure*}[!t]
\centering
\includegraphics[width=1\textwidth]{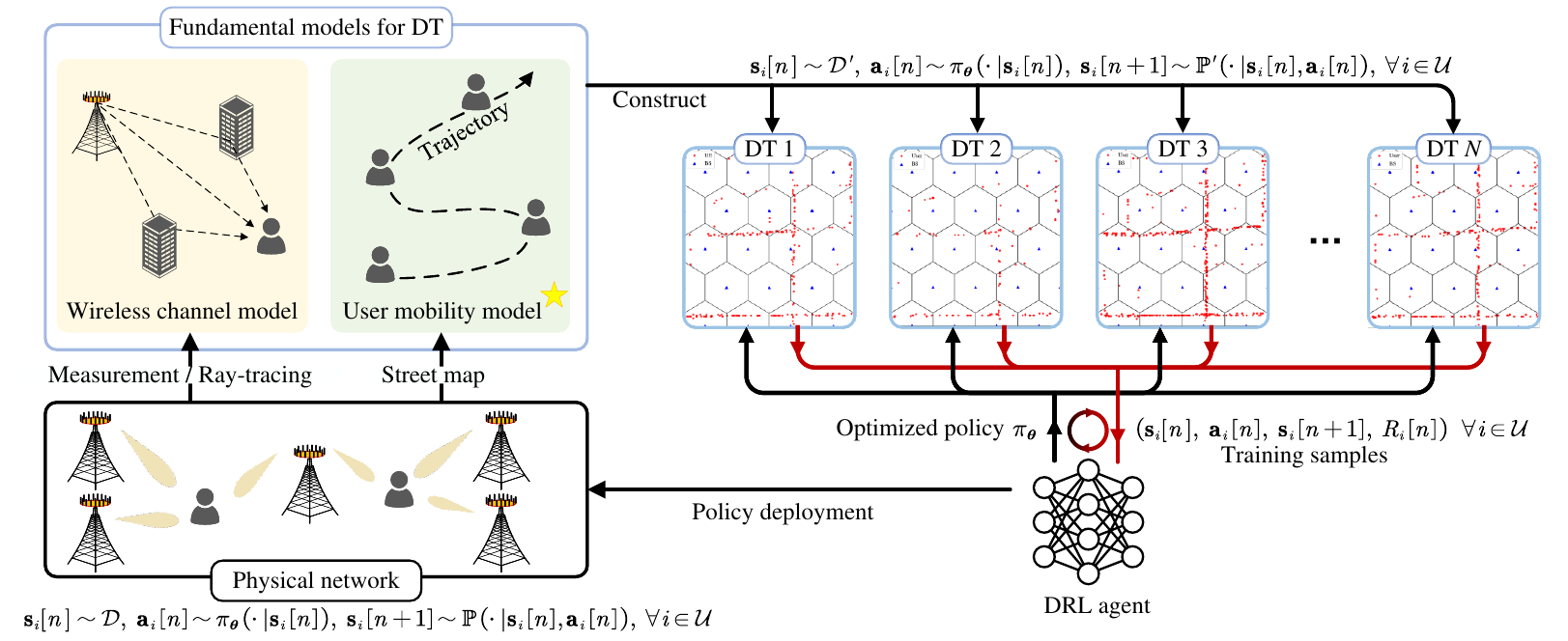}
\caption{Parallel DT framework.}
\label{Frame}
\end{figure*}

We now describe the entire architecture of the parallel DT framework for user association and load balancing in Fig.~\ref{Frame}. To devise the optimal user association policy in (\ref{maximize}), we first establish the fundamental models for DT, including a wireless channel model based on measurement and ray-tracing, and a user mobility model based on Map2Traj. Following the method outlined in Sec.~\ref{constructDT}, these models are used to create multiple DT environments, which maintain similar state distributions $\mathcal{D}^\prime$ and transition probability $\mathbb{P}^\prime$ as those in the physical wireless network. Next, the parameterized policy $\pi_{\boldsymbol{\theta}}$ (the DRL agents) is deployed into DTs to make association decisions. Each time DT environments take a step, $\sum_{k=1}^{N}|\mathcal{U}_k|$ training samples are generated, where $|\mathcal{U}_k|$ represents the number of users in the $k$-th DT. These samples are aggregated for batch training of $\pi_{\boldsymbol{\theta}}$ through gradient descent. The updated policy is then redeployed in the DT environments for further exploration. This process is repeated iteratively until the policy converges.

In summary, the proposed parallel DT-driven DRL method consists of the distributed DRL method, the LVM-enhanced DT, and the parallel DT framework. It addresses high trial-and-error costs and poor physical network performance by training in virtual DT environments. It also alleviates the strong correlation and non-stationarity in the real environment through parallelization, which therefore improves training efficiency and the generalization ability of DRL models. Note that the LVM-enhanced DT and parallel DT framework are versatile and applicable for other DT-driven network optimization tasks, particularly mobility-sensitive ones, such as wireless resource management \cite{9329087} and task offloading in mobile edge computing \cite{10024766}.

\section{Performance Evaluation}
In this section, we assess the efficacy of our proposed parallel DT-driven DRL method. First, we examine the established DT's fidelity to the real environment and its reliability in the DRL training. Then, a comprehensive evaluation of the proposed parallel DT-DRL method is conducted, along with a comparison against the DRL method trained directly in the real environment.

\subsection{Experiment Design}
In this paper, one of the main contributions is the proposed Map2Traj model for user mobility prediction. Since wireless channel modeling has been extensively studied, we first focus on the user mobility model to evaluate the DT's fidelity. We generate trajectories using various mobility models, including Map2Traj, existing random mobility models, and a state-of-the-art trajectory generation model. We then compare the generated trajectories to real ones based on two key factors for constructing DT as discussed in Sec.~\ref{constructDT}: \textit{trajectory patterns} and \textit{spatial distribution}.

After the similarity comparison on generated trajectories, we evaluate the reliability of the proposed Map2Traj-based DT in DRL training. We first establish an environment with real trajectories for deployment, along with several DT environments based on selected mobility models. To isolate the impacts of mobility models, all environments share the same wireless channel and maintain a constant number of users. We will regard the real-trajectory-based environment as the `real environment' throughout the experiment. DRL agents are initially trained in the DT environments, and once convergence is achieved, they are deployed in the real environment for performance evaluation. For comparison, we also include the performance of a DRL agent trained directly in the real environment as a benchmark for the optimal result.

To evaluate the parallel DT-DRL method, we train DRL agents using both a single DT and multiple parallel DT environments, comparing their performance in both DT and real environments. For a more comprehensive evaluation, we introduce $\pm5\%$ disturbance in the wireless channel of the real environment to simulate errors in the wireless channel modeling within the DT. Additionally, the number of users, $|\mathcal{U}|$, varies randomly between 100 and 400 in both DT and real environments to evaluate DRL performance under dynamic scenarios.

\subsection{Dataset and Experiment setting}

Map2Traj is trained using a real-world vehicle trajectory dataset from Xi'an, China, recorded in 2016 \cite{didi2017gaia}, alongside the OpenStreetMap dataset of the same year. These datasets are sourced from the ChinaGEOSS Data Sharing Network. The trajectories are all within the latitude range of 34.21 to 34.28 and the longitude range of 108.912 to 108.996. To avoid data leakage in the zero-shot trajectory generation, the training set is limited to longitudes between 108.912 and 108.974, while the test set extends from 108.974 to 108.996. 
Notice that the correlation between the street map and trajectory, as shown in Fig.~\ref{correlation}, remains consistent under transformations such as rotation and reflection. This inherent property can be leveraged for data augmentation. We randomly rotate and flip both street maps and trajectory data in training to enhance the generalization capability of the Map2Traj model.

The mobility models employed for comparison are mainly random mobility models, typically the classic RWP and GM models. Moreover, we develop their variants to adapt to geographical constraints. The variant of RWP is the map-restricted random waypoint (M-RWP), where the destinations are confined within the street map area. The trajectory between points is determined by a breadth-first search (BFS) algorithm to ensure that the shortest path remains within streets. Similarly, the map-restricted Gauss Markov model (M-GM) is also devised to restrict user movements within street areas. In addition, we include the state-of-the-art trajectory generation model DiffTraj \cite{DiffTraj}, despite its lack of zero-shot trajectory generation capability, in order to benchmark the generation quality of our proposed model. It is important to note that DiffTraj is trained on the complete trajectory dataset \cite{didi2017gaia}, including those from the test area. The relevant data is sourced from the DiffTraj-generated synthetic dataset, SynMob, provided by the authors of DiffTraj in \cite{SynMob}.

\begin{figure}[!t]
\centering
\includegraphics[width=0.2\textwidth]{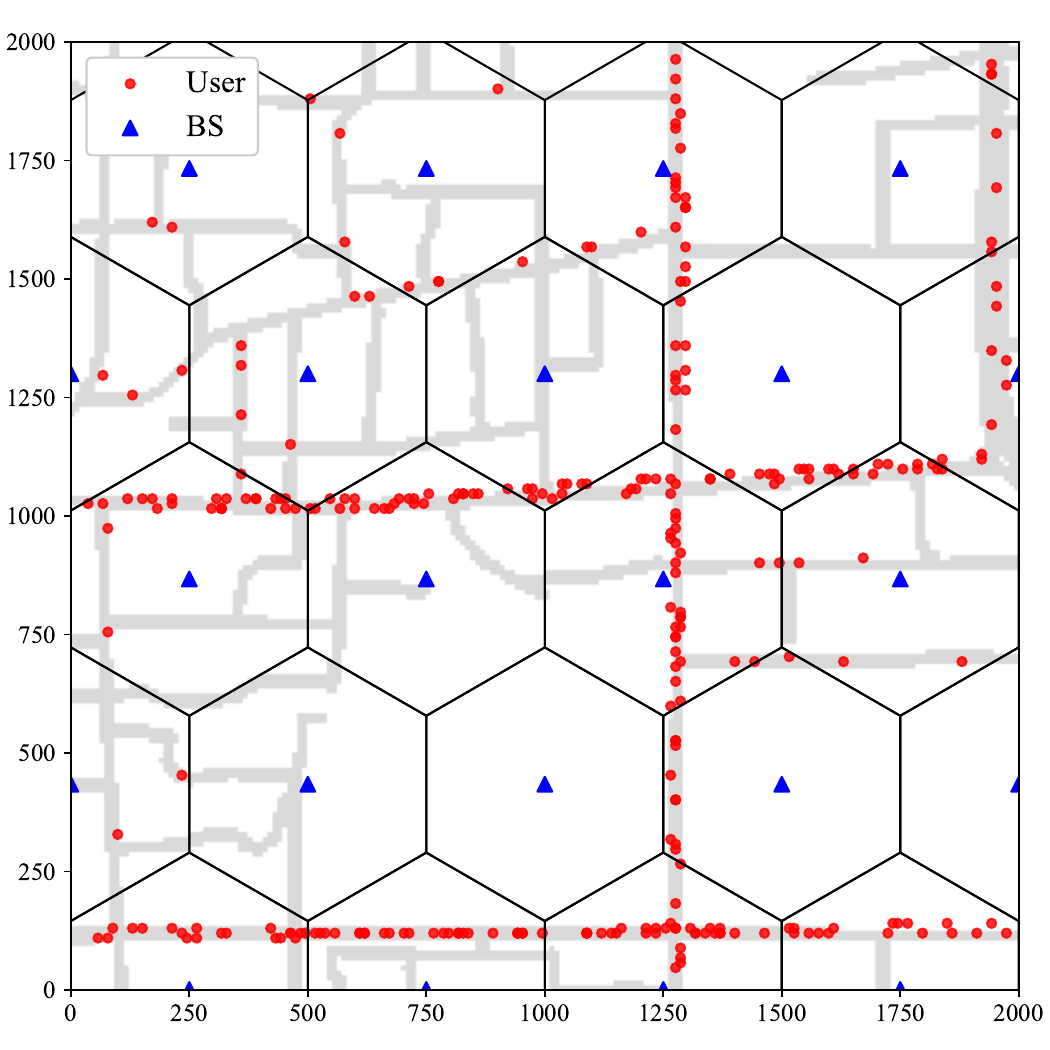}
\caption{Experiment scenario.}
\label{Fig4}
\end{figure}

In terms of the wireless network, we consider a typical 2km$\times$2km urban area where 22 BSs are densely deployed in a hexagonal pattern, with a 500 m interval. Each BS possesses multi-band capabilities, supporting connections at 3.7 GHz with a 40 MHz bandwidth and 0.7 GHz with a 10 MHz bandwidth. The total number of BSs is $|\mathcal{B}|=22\times2=44$. Within this area, users continuously move and connect to BSs based on a specific user association policy, as illustrated in Fig.~\ref{Fig4}. The wireless channel in this environment employs the urban macrocell pathloss model from 3GPP \cite{3gpp}. In addition, shadow fading is implemented through the sum-of-sinusoids method to ensure spatial consistency, which has been used in QuaDRiGa \cite{quadriga_website}. Handover events are modeled following an existing work \cite{10298039}. Specifically, the handover interruption time $T_{\mathrm{HO},i}$ is characterized as a discrete random variable. The handover process is anticipated to succeed with an 80\% probability, incurring an interruption of 20 ms. Conversely, with a 20\% probability, the handover fails, leading to a more significant interruption duration of 90.76 ms. The total time slot duration $T_s$ is 100 ms. The parameters for the wireless environment are listed in Table~\ref{para}.

\begin{table}[t]
\centering
\caption{Wireless Simulation Parameters}
\begin{tabular}{|c|c|}
\hline
\textbf{Transmit power}          & 46 dBm                                     \\ \hline
\textbf{Frequency bands}         & [3.7, 0.7] GHz                       \\ \hline
\textbf{Bandwidth}               & 3.7 GHz: 40 MHz  \ \ 0.7 GHz: 10 MHz \\ \hline
\textbf{Pathloss and shadowing}  & Urban macrocell model from 3GPP \cite{3gpp}   \\ \hline
\textbf{Number of BS}            & 22 (location) $\times$ 2 (frequency) = 44                              \\ \hline
\textbf{Number of users}         & 100 -- 400                                       \\ \hline
\textbf{Inter-site distance}   & 500 m                                      \\ \hline
\textbf{Noise}                   & -174 dBm/Hz                                \\ \hline
\textbf{BS height}               & 25 m                                       \\ \hline
\textbf{User height}             & 1.5 m                                      \\ \hline
\end{tabular}
\label{para}
\end{table}

\subsection{Evaluation Metrics}
The metrics we employed for the evaluation on \textit{trajectory pattern} include Edit Distance on Real Sequences (EDR) \cite{editdistance} and Dynamic Time Wrapping (DTW)  \cite{dtw}, which are widely used in mobility analysis \cite{gis}. EDR quantifies the minimum number of operations required to make two trajectories match, where a match is defined when the distance between corresponding points is less than a threshold of $\tau$ meters, and $\tau = 20$ in this experiment. DTW calculates the squared Euclidean distance between two trajectories through a dynamic programming alignment algorithm.

The metrics for \textit{spatial distribution} consist of cosine similarity and Wasserstein distance \cite{ruschendorf1985wasserstein}. Cosine similarity is a widely used measure of similarity between two vectors. While it reflects the similarity between probability distributions, it falls short in expressing the spatial correlation between adjacent blocks in two-dimensional (2D) distributions. To address this limitation, we introduce the Wasserstein distance, which quantifies the minimal cost required to transform one probability distribution into another through an optimal transport plan. In our experiment, the Wasserstein distance measures the effort required to transform the spatial distribution of generated trajectories into that of real trajectories. However, computing Wasserstein distance for 2D distributions entails solving a complex high-dimensional linear programming problem, which is computationally impractical for 192$\times$192 arrays. Consequently, we use the sliced Wasserstein distance \cite{kolouri2019generalized} as an alternative.

The performance metrics for user association and load balancing encompass the 5th percentile user rate (5\% rate) to evaluate cell-edge performance, and the logarithmic mean of all user rates to indicate overall network utility. The latter also serves as the global reward for the DRL agents.

\subsection{Numerical Results}
\begin{figure}[t!]
\centering
\subfloat[RWP]{\includegraphics[width=0.14\textwidth]{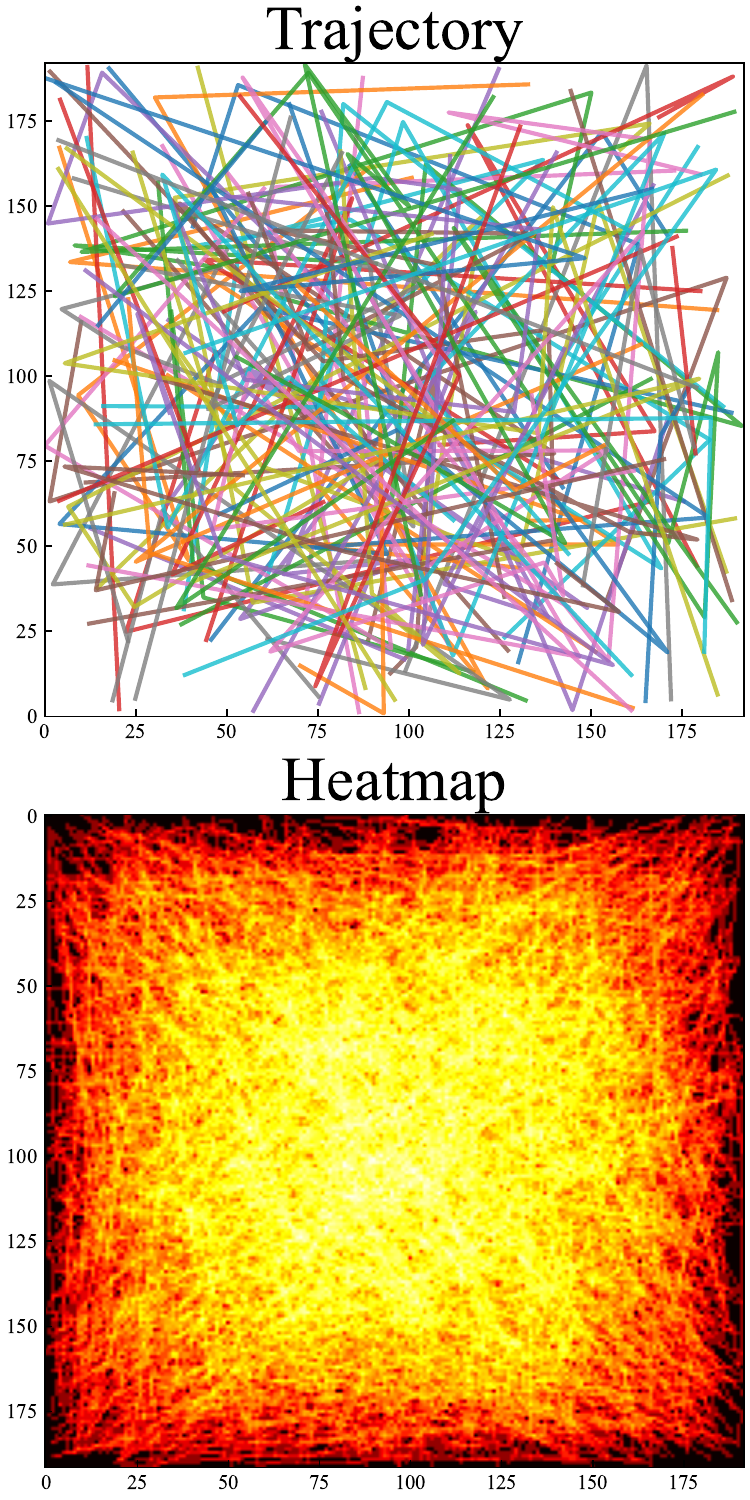}}
\subfloat[GM]{\includegraphics[width=0.14\textwidth]{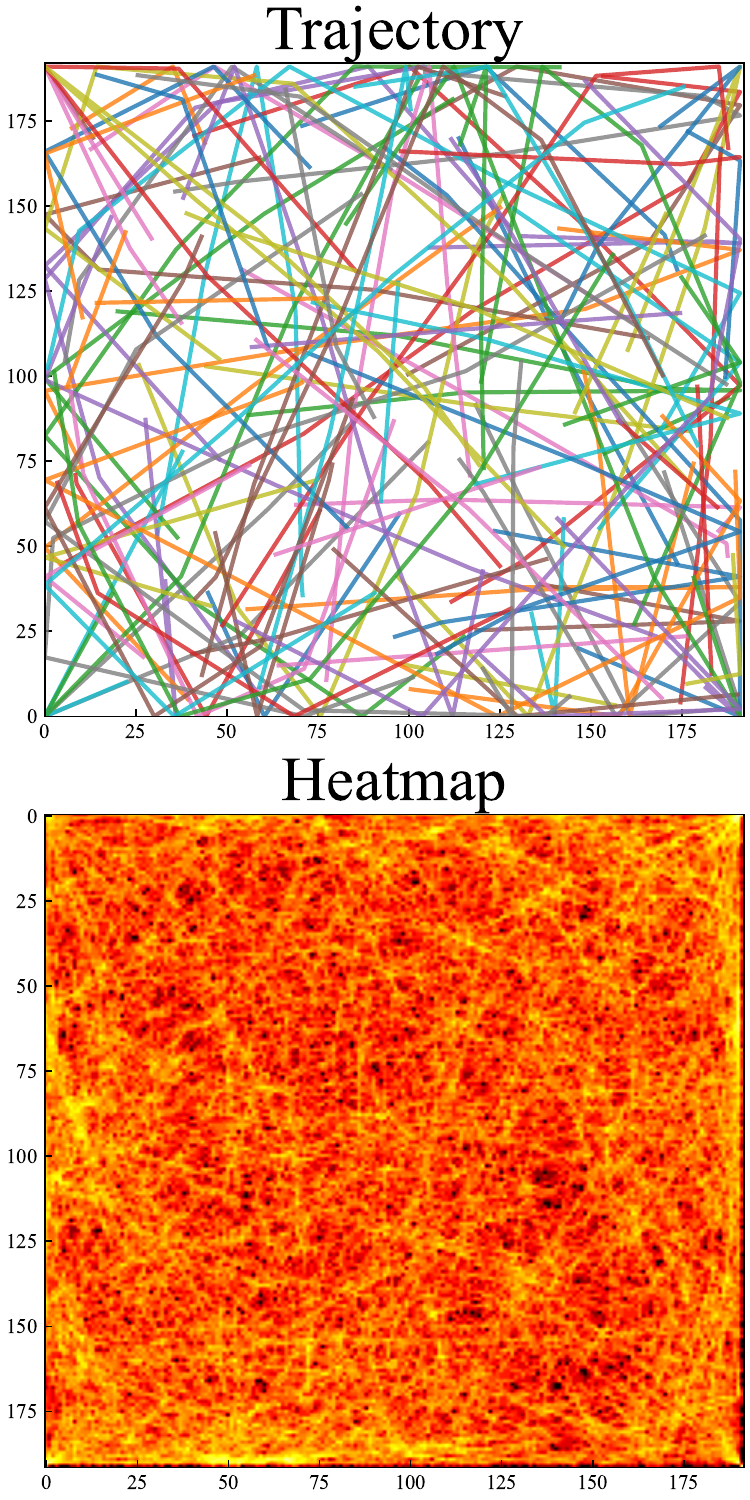}}
\subfloat[M-RWP]{\includegraphics[width=0.14\textwidth]{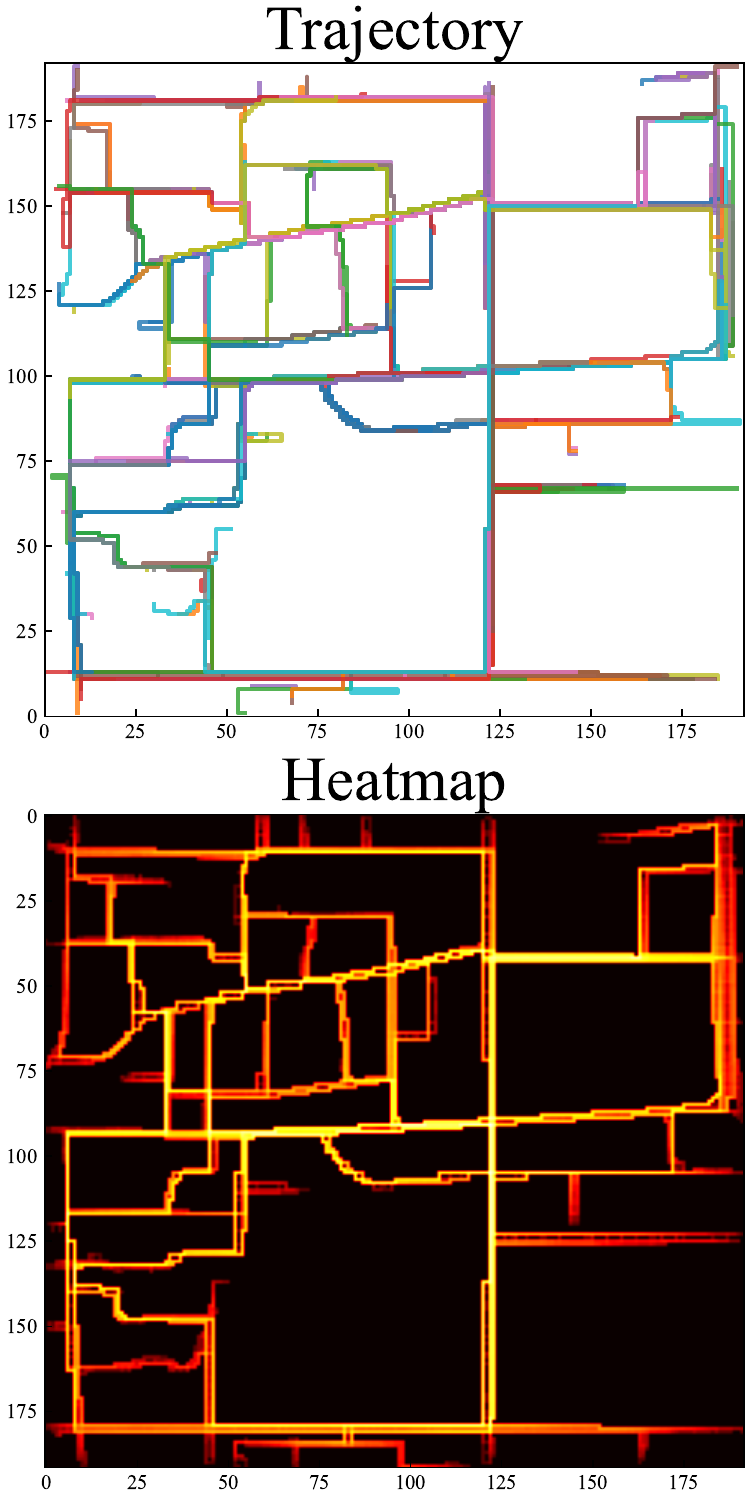}}

\subfloat[M-GM]{\includegraphics[width=0.14\textwidth]{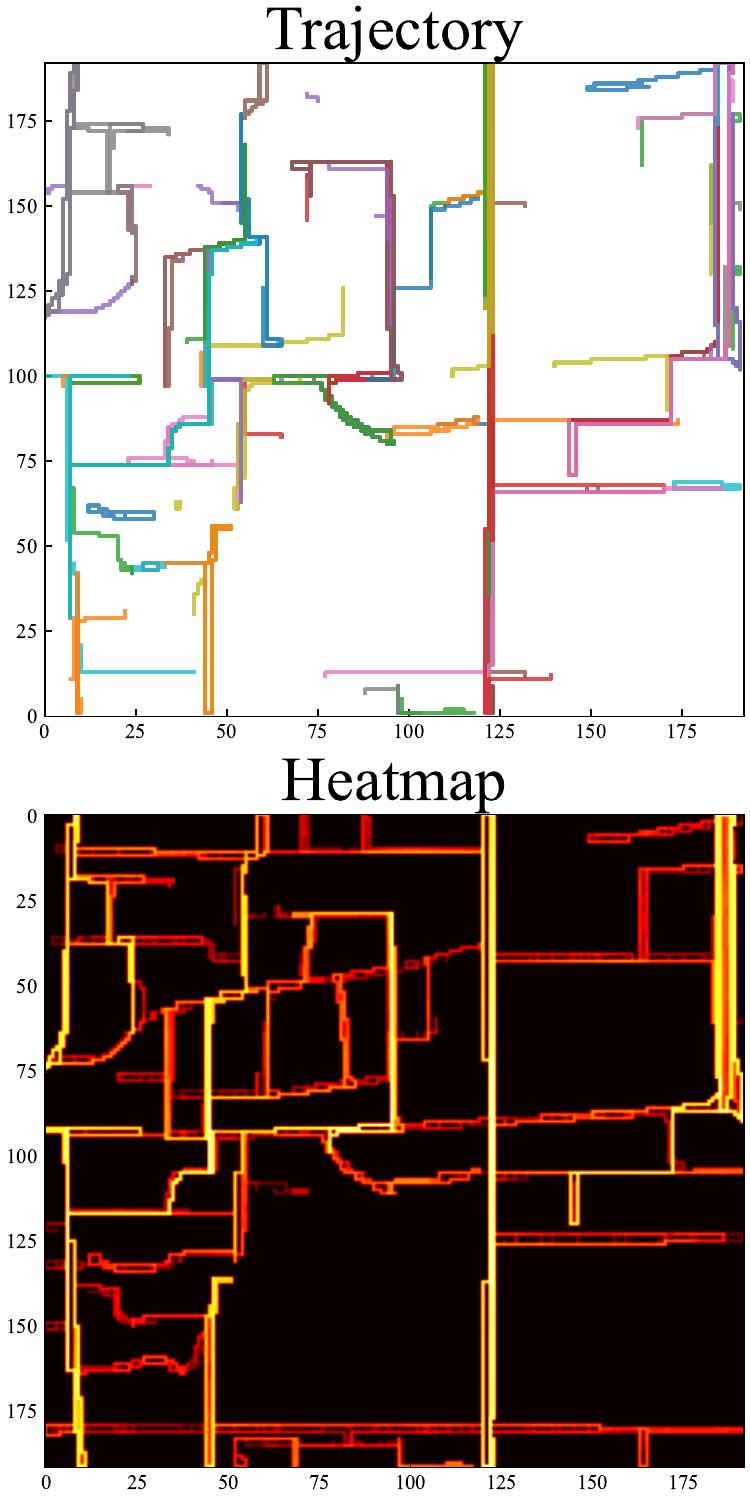}}
\subfloat[DiffTraj]{\includegraphics[width=0.14\textwidth]{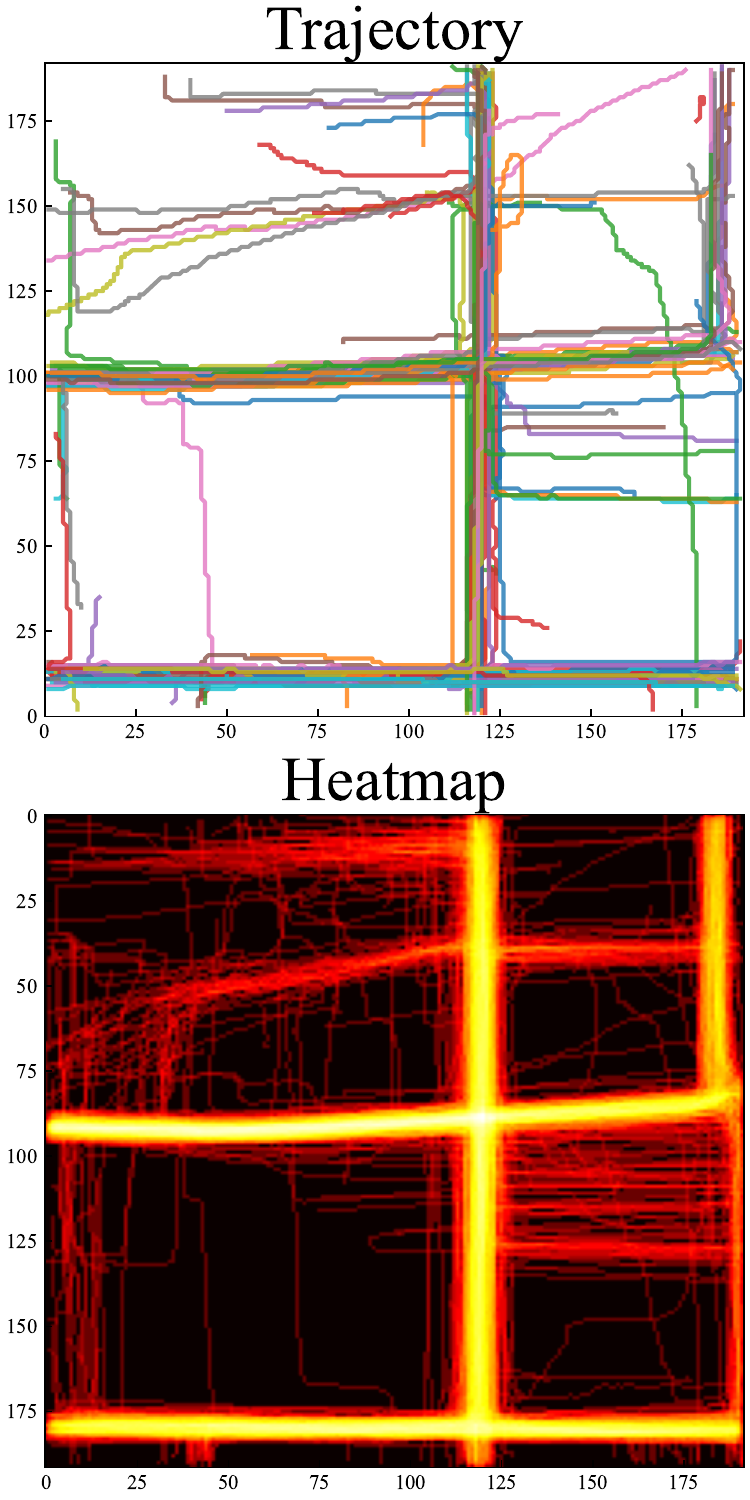}}
\subfloat[Map2Traj]{\includegraphics[width=0.14\textwidth]{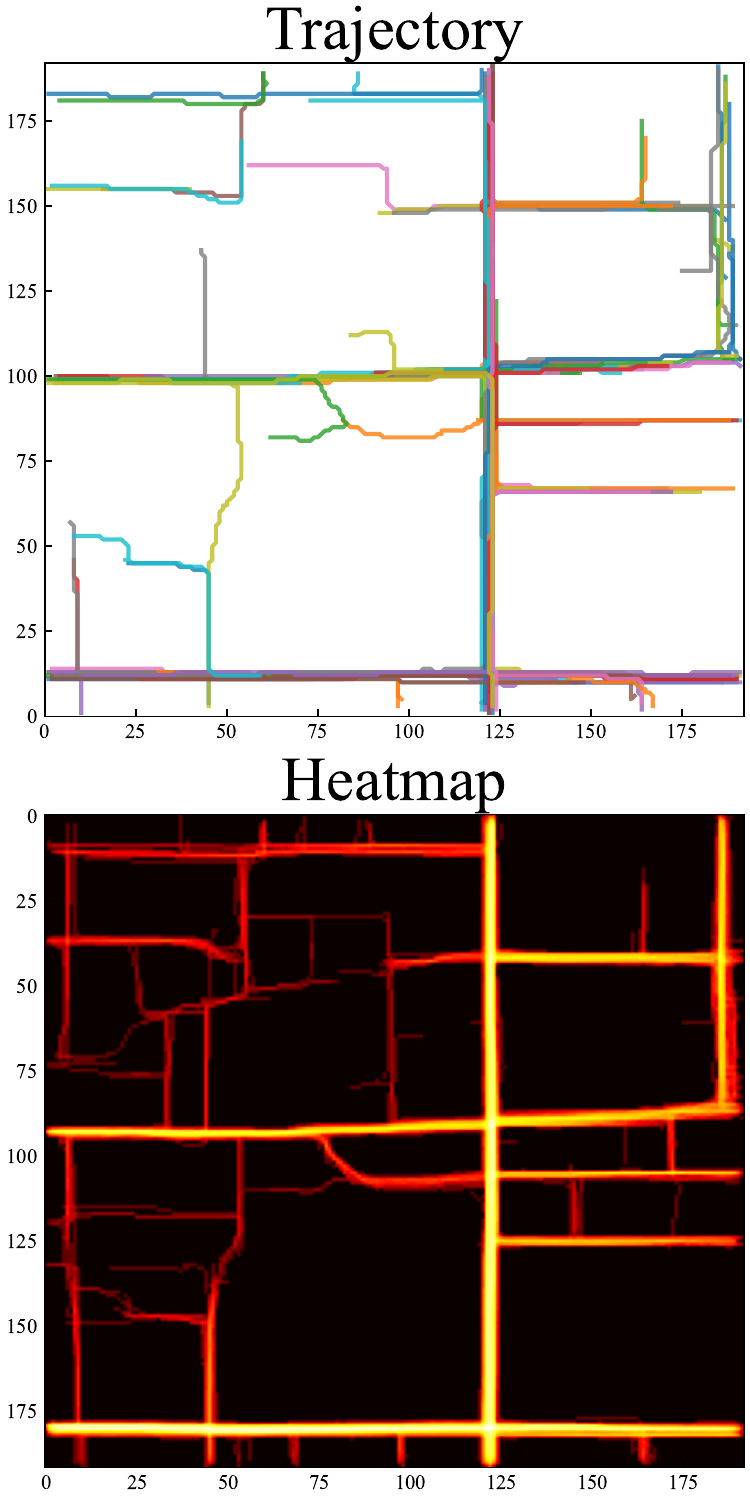}}

\caption{Generated trajectories and heatmaps.} 
\label{Fig3}
\end{figure}

We select the area mentioned in \ref{constructDT} and Fig.~\ref{correlation} as the test area. Fig.~\ref{Fig3} displays the generated trajectories by all methods alongside corresponding heatmaps. Traditional random mobility models, i.e., RWP and GM, result in chaotic trajectories and heatmaps that bear no resemblance to real-world patterns. While map-restricted models, M-RWP and M-GM, show some similarity in trajectory patterns to real ones, they still fall short in spatial distribution similarity due to the absence of a learning mechanism. As expected, DiffTraj demonstrates high similarity to real trajectories and heatmaps because it is trained directly on real data. In contrast, our proposed Map2Traj, even in a zero-shot scenario, produces results comparable to real trajectories and even surpasses DiffTraj in trajectory details.

\begin{table}[t]
    \centering
    \caption{Quantified Evaluation of Trajectory Generation Performance}
    \begin{tabular}{ccccc}
        \toprule \multirow{2.5}{*}{\textbf{Mobility model}}  & \multicolumn{2}{c}{\textbf{Trajectory pattern}} & \multicolumn{2}{c}{\textbf{Spatial distribution}}
        \\
        \cmidrule(r){2-3}\cmidrule(r){4-5}
        &EDR $\downarrow$ & \ \ DTW $\downarrow$ \ & CS $\uparrow$ & WD $\downarrow$\\
        \midrule 
        RWP & 264.1 & 76.74 & 0.1537 & 21.01\\
        GM &213.1& 82.96  & 0.1698 & 19.21\\
        M-RWP & 192.0 & 21.44 & 0.3081 & 22.91\\
        M-GM & 155.4 & 33.57  & 0.2793 & 26.32\\
        DiffTraj & 68.35 & 13.63 & 0.5573 & 9.134\\
        \textbf{Map2Traj} &\textbf{21.47} & \textbf{8.933}& \textbf{0.6834} & \textbf{6.096}\\
        \midrule 
        Real trajectories & 7.570 & 1.018 & 0.9959 &  2.569\\
        \bottomrule
    \end{tabular}
    \begin{tablenotes}
    \item[] \textcolor{black}{CS and WD stand for cosine similarity and Wasserstein distance.}
    \end{tablenotes}
    \label{tab2}
\end{table}

Further, we quantitatively evaluate the similarity of the generated trajectories with real ones using the above-mentioned metrics. Considering the stochastic nature of trajectory generation, we generate 1,000 trajectories for each method and compare them against a benchmark of 1,000 real trajectories. For each generated trajectory, we calculate EDR and DTW with real ones, identifying the minimum value as the representative metric. This process is repeated for all 1,000 trajectories to determine the average \textit{trajectory pattern} similarity between the generated set and the real dataset. For \textit{spatial distribution} similarity, we sum all 1,000 trajectories into a grayscale image, normalize them to represent the probability of user presence at various locations, and calculate the cosine similarity and Wasserstein distance between the generated user distribution and the real one. We also calculate metrics for another set of 1,000 real trajectories to serve as optimal similarities. Table~\ref{tab2} presents the quantified similarity comparison among different trajectory sets. The results indicate that our proposed Map2Traj significantly exceeds the random mobility models and map-restricted ones, producing synthetic trajectories closely resembling real ones in both \textit{trajectory pattern} and \textit{spatial distribution}. This suggests that Map2Traj has effectively learned the correlation between street maps and actual trajectories, and can be employed in the construction of a faithful DT.

It is encouraging to see that our zero-shot Map2Traj model outperforms the area-specific DiffTraj. The advantage of Map2Traj lies in the continuous guidance from street maps throughout the denoising process, while DiffTraj only involves some trip information as a condition. On the other hand, DiffTraj fixes the trajectories into uniform shapes through sampling, instead of transforming them into images, potentially leading to information loss. However, it is crucial to acknowledge that existing learning-based methods, including DiffTraj, have the potential to outmatch Map2Traj given a sufficiently large training dataset, a wider and deeper network structure, or in some specific test scenarios. The primary contribution of Map2Traj is the zero-shot trajectory generation ability, rather than merely surpassing existing area-specific trajectory generation techniques.

\begin{figure}[!t]
\centering
\includegraphics[width=0.4\textwidth]{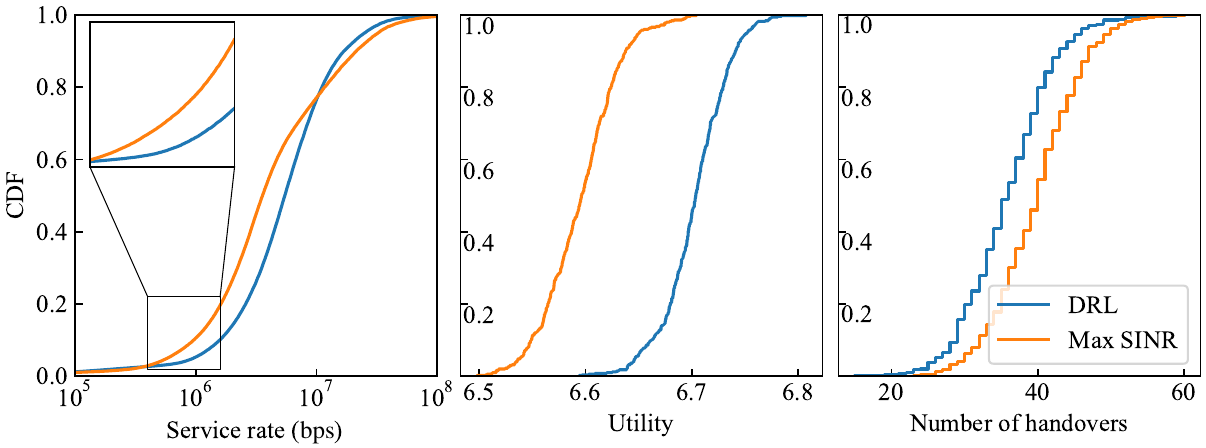}
\caption{DRL performance in the training environment.}
\label{dtenv}
\end{figure}

Next, we construct various DT environments based on different user mobility models and train DRL agents through the method outlined in Section.~\ref{DRLmethod}. Taking Map2Traj-based DT environment as an example, Fig.~\ref{dtenv} presents a comparative analysis of the cumulative density function (CDF) for service rate, overall network utility, and handover count between the converged DRL method and a traditional Max SINR method. By marginally sacrificing the service rate for the top 20\% of users, the DRL strategy achieves a significant enhancement in service rates for the remainder, especially the edge users in the lower 20\%. This indicates that DRL effectively balances load within the wireless network. It also reduces the number of handovers and eventually improves the overall network utility. The DRL agents trained in other DT environments exhibit similar superiority in enhancing cell-edge performance and overall network utility. As shown in Table~\ref{tab3}, the DRL methods notably surpass the traditional Max SINR approach in all DT training environments. 

\begin{figure}[!t]
\centering
\includegraphics[width=0.5\textwidth]{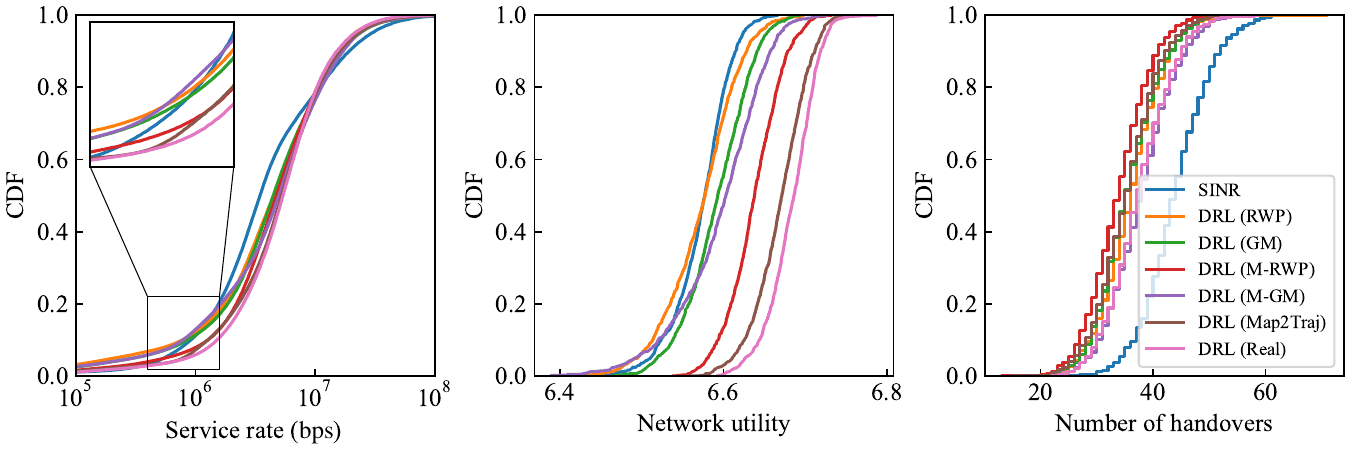}
\caption{DRL performance after deployment in a real environment.}
\label{Fig5}
\end{figure}

\begin{table}[t]
    \centering
    \caption{User Association Performance in Training Environments}
    \renewcommand{\arraystretch}{0.7}
    \begin{tabular}{cccc}
        \toprule
        \textbf{Mobility model} & \textbf{Method} & \textbf{$5\%$ rate ($\times 10^5$)} $\uparrow$ & \textbf{Utility} $\uparrow$\\
        \midrule 
        \multirow{2}{*}{RWP} & Max SINR & 5.030 & 6.561  \\
        & DRL & 7.241 & 6.672 \\
        \midrule
        \multirow{2}{*}{GM} & Max SINR & 7.662 & 6.721\\
        & DRL & 8.881 & 6.788  \\
        \midrule
        \multirow{2}{*}{M-RWP} & Max SINR & 5.250 & 6.603  \\
        & DRL & 9.752 & 6.707 \\
        \midrule
        \multirow{2}{*}{M-GM} & Max SINR & 5.948 & 6.674\\
        & DRL & 11.373 & 6.766  \\
        \midrule
        \multirow{2}{*}{Map2Traj} & Max SINR & 5.871 & 6.599 \\
        & DRL & 9.877 & 6.705 \\
        \midrule
    \end{tabular}
    \label{tab3}
\end{table}

However, what truly matters for DT-driven DRL methods is not the training performance in DT, but their efficacy in the real environment. We subsequently deploy these trained agents into the environment with real trajectories, and the results are presented in Table~\ref{tab4}. The numerical results confirm that agents trained with random mobility models exhibit severe performance degradation, in some cases deteriorating to levels comparable to or even worse than the Max SINR method. In comparison, the DRL agent trained using the Map2Traj-based mobility model maintains competitive performance, achieving 91.6\% of the cell-edge user performance and 96.6\% of the network utility compared to agents trained in real environments. The 96.6\% network utility is derived from the ratio $10^{6.669}/10^{6.684}$, given that network utility is a logarithmic metric.

\begin{table}[t]
    \centering
     \caption{User Association Performance in Deployment}
     \renewcommand{\arraystretch}{0.7}
    \begin{tabular}{ccc}
        \toprule
        \textbf{Method} & \textbf{$5\%$ rate ($\times 10^5$)} $\uparrow$ & \textbf{Utility} $\uparrow$\\
        \midrule
         Max SINR & 5.408 & 6.575\\
         DRL (RWP) & 2.238 & 6.576 \\
         DRL (GM) & 2.771 &  6.587\\
         DRL (M-RWP) & 5.568 & 6.637 \\
         DRL (M-GM) & 3.145 &  6.601\\
         DRL (Map2Traj) & \textbf{7.823} & \textbf{6.669}\\
        \midrule
         DRL (Real) & 8.538 & 6.684 \\

        \bottomrule
    \end{tabular}
   
    \label{tab4}
\end{table}

To provide a more nuanced view of the performance, we present the CDF for each method in Fig.~\ref{Fig5}. Although all DRL methods successfully reduce the number of handover events, those trained with random mobility models struggle to guarantee service rates for edge users. Methods such as RWP, GM, and M-GM underperform the traditional Max SINR method in terms of service rate for edge users in the bottom 10\%. In contrast, the Map2Traj-based DRL agent maintains superior load-balancing performance. All these results demonstrate that Map2Traj-based DT has efficacy comparable to the real environment in training DRL agents.

We also investigate the relationship between the fidelity of user mobility models and the performance of correspondingly trained DRL agents. The fidelity is quantified by the cosine similarity and the negative logarithm of DTW. For better comparison, we normalize the 5\% user rate and the overall network utility into the range $\left[0,1\right]$.
\begin{figure}[!t]
\centering
\includegraphics[width=0.35\textwidth]{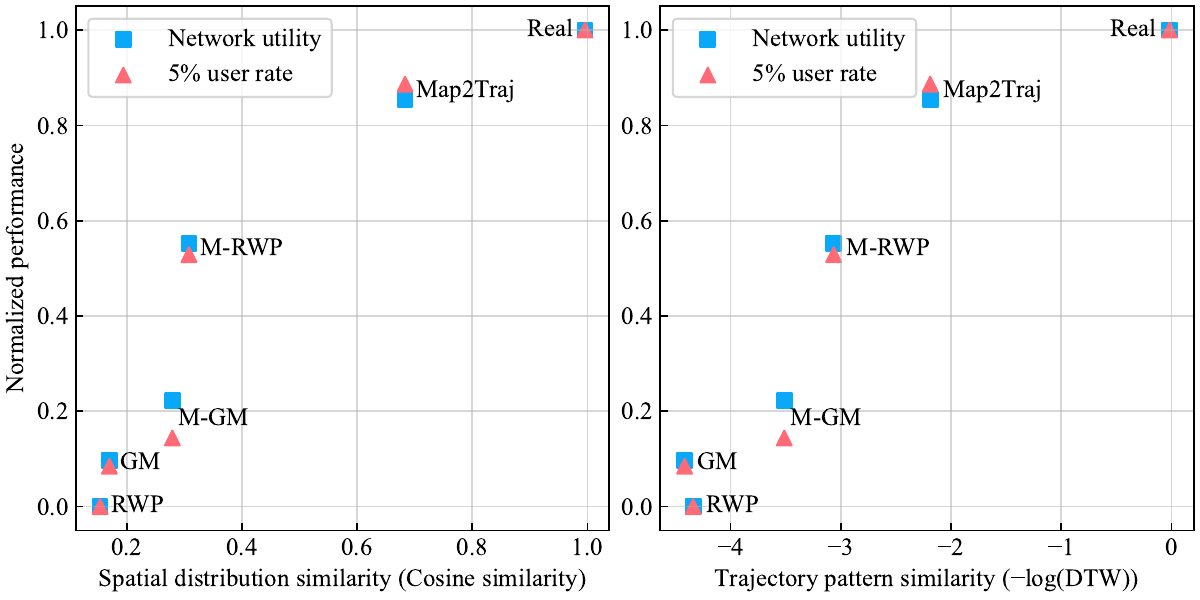}
\caption{Correlation between user mobility fidelity and DRL performance.}
\label{Fig6}
\end{figure}
As illustrated in Fig.~\ref{Fig6}, both similarities in \textit{spatial distribution} and \textit{trajectory pattern} are positively correlated with the performance of DRL agents. This finding verifies our analysis of the prerequisite for constructing faithful DT for network optimization, as discussed in Section~\ref{constructDT}.

\begin{figure}[!t]
\centering
\includegraphics[width=0.4\textwidth]{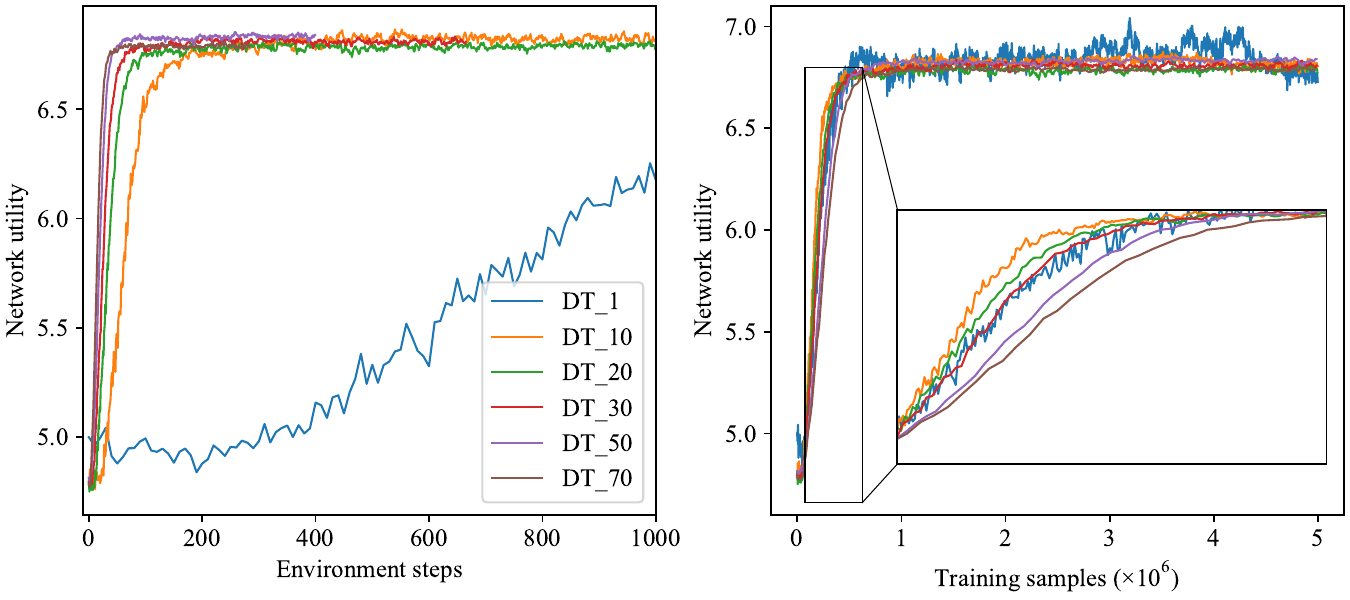}
\caption{Network utility in DRL training process}
\label{rewardcompare}
\end{figure}

For a thorough evaluation of the proposed parallel DT-DRL method, we train DRL agents across a range of parallel DT environments with varying user counts and wireless channel disturbances. To ensure fairness, the number of training samples is capped at $5\times10^6$, with the total environment step count being contingent upon the number of parallel DTs utilized. For instance, with an average user count of $250$, the step count for a single DT environment (DT\_1) is set to $5\times10^6/250=20,000$, whereas 10 parallel DT environments (DT\_10) have $2,000$ steps each. Fig.~\ref{rewardcompare} shows the evolution of network utility, averaged per environment step, throughout the training process. In the single training environment DT\_1, network utility fluctuates during training due to changes in user numbers, making it harder for DRL agents to explore globally optimal policies. Introducing parallel DTs helps stabilize these fluctuations and accelerates convergence. However, as shown on the right side of Fig.~\ref{rewardcompare}, the increase in the parallel DTs can diminish the training benefit derived from each sample.

\begin{figure}[!t]
\centering
\includegraphics[width=0.35\textwidth]{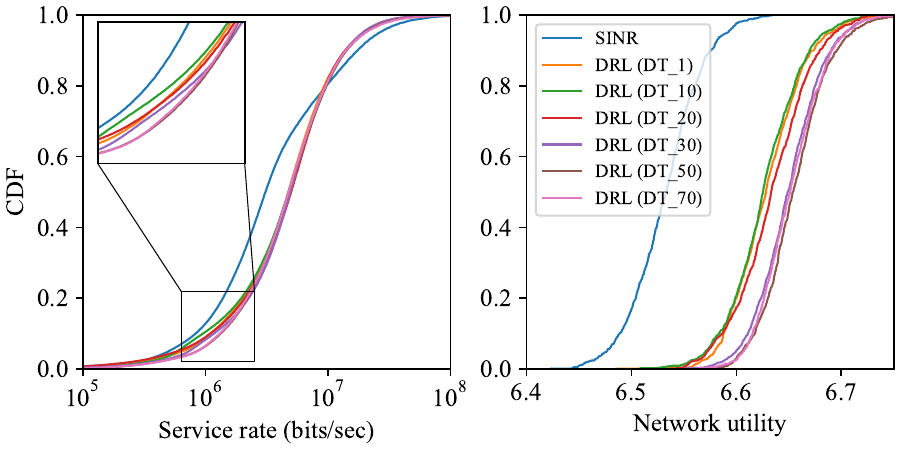}
\caption{DRL performance in the training environment}
\label{DTpara}
\end{figure}

We first evaluate the trained DRL agents in another DT environment, where the number of users is randomly initialized between 100 and 400 and varies over time. Figure~\ref{DTpara} shows that as the number of parallel DT environments in training increases, both edge user performance and network utility improve, peaking at 50 parallel environments. This outcome supports our assertion on the limitations of single-environment training in Section~\ref{parallelframework}. However, beyond this point, further increasing parallelism leads to performance decline, as illustrated by DT\_70, due to the fixed total number of samples and reduced sample efficiency.

\begin{figure}[!t]
\centering
\includegraphics[width=0.35\textwidth]{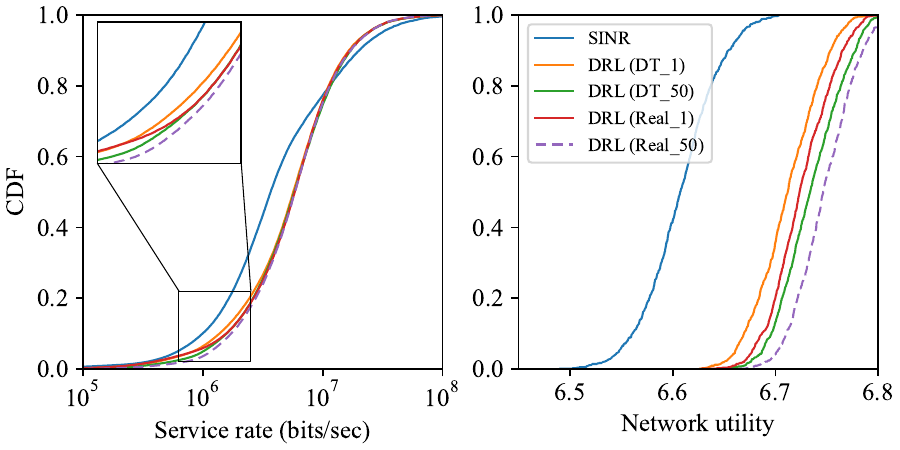}
\caption{DRL performance in the real environment}
\label{Realparal}
\end{figure}

Finally, we test the agents in the real environment. We compared agents trained in a single DT environment (DT\_1) and the best-performing parallel environment (DT\_50) against those trained directly in a real-world single environment. We also created a parallel real environment as a benchmark, though this is impractical in reality due to the uncontrollable and non-reproducible nature of real environments. As illustrated in Fig.~\ref{Realparal} and Table~\ref{tab5}, the parallel DT-DRL method outperforms the DRL method trained in a single real-world environment, achieving 118.8\% cell-edge user performance and 102.8\% network utility. Notably, this is accomplished without any interaction with the physical wireless network during training.

\begin{table}[t]
    \centering
     \caption{User Association Performance in Deployment}
     \renewcommand{\arraystretch}{0.7}
    \begin{tabular}{ccc}
        \toprule
        \textbf{Method} & \textbf{$5\%$ rate ($\times 10^5$)} $\uparrow$ & \textbf{Utility} $\uparrow$\\
        \midrule
         Max SINR & 6.171 & 6.527\\
        DRL (DT\_1) & 8.324 & 6.625 \\
         DRL (DT\_50) & \textbf{10.053} & \textbf{6.652} \\
         DRL (Real\_1) & 8.462 & 6.640 \\
         \midrule
         DRL (Real\_50) & 11.683 & 6.667 \\
        \bottomrule
    \end{tabular}
    \begin{tablenotes}
    \item[] \textcolor{black}{\qquad \quad  (Real\_50) is unachievable in the real world}
    \end{tablenotes}
    \label{tab5}
\end{table}

\section{Conclusion}
In this paper, we have proposed a parallel DT-driven DRL method designed for user association and load balancing in dynamic wireless networks. To accommodate varying user numbers, we have introduced a distributed DRL strategy with an enhanced PPO network structure to speed up convergence. To address high trial-and-error costs in real-world training and poor physical network performance before convergence, we have developed an LVM-enhanced DT for wireless networks, featuring a zero-shot generative user mobility model named Map2Traj. DRL agents undergo training within this DT environment, thereby avoiding direct interaction with the physical network. Furthermore, a parallel DT framework has been developed to alleviate the strong correlation and non-stationarity in the real environment through parallelization, which therefore improves training efficiency and the generalization ability of DRL models. The proposed parallel DT-driven DRL method is versatile and applicable for other DT-driven network optimization tasks, such as wireless resource management and task offloading in mobile edge computing.



\ifCLASSOPTIONcaptionsoff
  \newpage
\fi



\footnotesize
\bibliographystyle{IEEEtran}

\bibliography{IEEEabrv,IEEEexample}
\end{document}